\pdfoutput=1
\documentclass{article}

\usepackage[table]{xcolor}
\usepackage[main, final]{neurips_2026}
\usepackage{mwe}
\usepackage{float}
\usepackage{multirow}
\usepackage[table]{xcolor}
\usepackage{booktabs}
\usepackage{arydshln}
\usepackage{adjustbox}
\usepackage{graphicx} 
\usepackage{wrapfig}

\usepackage{caption}
\usepackage[labelformat=parens]{subcaption}

\usepackage{listings}
\usepackage{xcolor}

\usepackage[utf8]{inputenc} 
\usepackage[T1]{fontenc}    
\usepackage{hyperref}       
\usepackage{url}            
\usepackage{booktabs}       
\usepackage{amsfonts}       
\usepackage{nicefrac}       
\usepackage{microtype}      
\usepackage{xcolor}         
\usepackage{amsmath}
\usepackage{cleveref}

\title{Learning Hierarchical Patch Splitting Policies for Faster Vision Transformers}

\author{%
  {\bfseries
  Aditya Gupta$^{1}$ \hspace{1.5em}
  Jean-Sebastien Dandurand$^{1}$ \hspace{1.5em}
  Kai Qiu$^{1}$ \hspace{1.5em}
  Rohan Choudhury$^{1}$} \\[0.5ex]
  {\bfseries L\'aszl\'o A.~Jeni$^{1}$} \\[0.9ex]
  \normalfont
  $^{1}$Carnegie Mellon University \\[0.6ex]
}

\begin{document}

\maketitle

\begin{abstract}
Vision Transformers (ViTs) are bottlenecked by the length of their input token
sequence. Standard patchification fixes this length with a single patch size applied across the image, regardless of regional content. Prior work on adaptive patch sizing reduces token count by assigning
coarser patches to visually simple regions, but relies on noisy pixel-level heuristics
such as local entropy to make patchification decisions.
We argue that this compute allocation should be grounded in the model's interpretation of task relevant information,
not in raw pixel statistics. We introduce \textbf{S}plit \textbf{P}olicy \textbf{L}earned \textbf{I}mage \textbf{T}okenization which learns to allocate token resolution based on what the model finds meaningful, producing shorter token sequences that are better
matched to task-relevant content.
 \textbf{SPLIT} matches the accuracy of
full-resolution ViT classification baselines while using 30\% fewer tokens.
The same gains hold on dense prediction: \textbf{SPLIT} achieves
competitive accuracy on COCO and
ADE20K, demonstrating that 
semantically aligned
patchification generalises well beyond classification.
\end{abstract}

\section{Introduction}
\label{sec:intro}

Vision Transformers (ViTs)~\citep{dosovitskiy2021vit} have become the
dominant architecture for various visual tasks, demonstrating strong
scalability and effectiveness across a wide range of
applications. However, their efficiency remains a major
challenge: self-attention scales quadratically with sequence length, and
because patch embedding uses a uniform grid, sequence length itself grows
quadratically with input resolution. As a result, doubling the resolution
of an image quadruples its token count and increases self-attention cost
by roughly $16\times$, making high-resolution inputs prohibitively
expensive to process.

To reduce the number of tokens processed while preserving task-relevant
information, many prior works reduce the token count after the tokens have
been fully computed. These methods typically leverage signals produced by
the Vision Transformer itself, such as self-attention scores \citep{liang2022evit} or
intermediate feature representations \citep{bolya2023tome}, to filter, merge, or refine
tokens~\citep{rao2021dynamicvit}, aligning token allocation with the model's learned
semantics.

\begin{figure}[ht]
\begin{center}
\includegraphics[width=\textwidth]{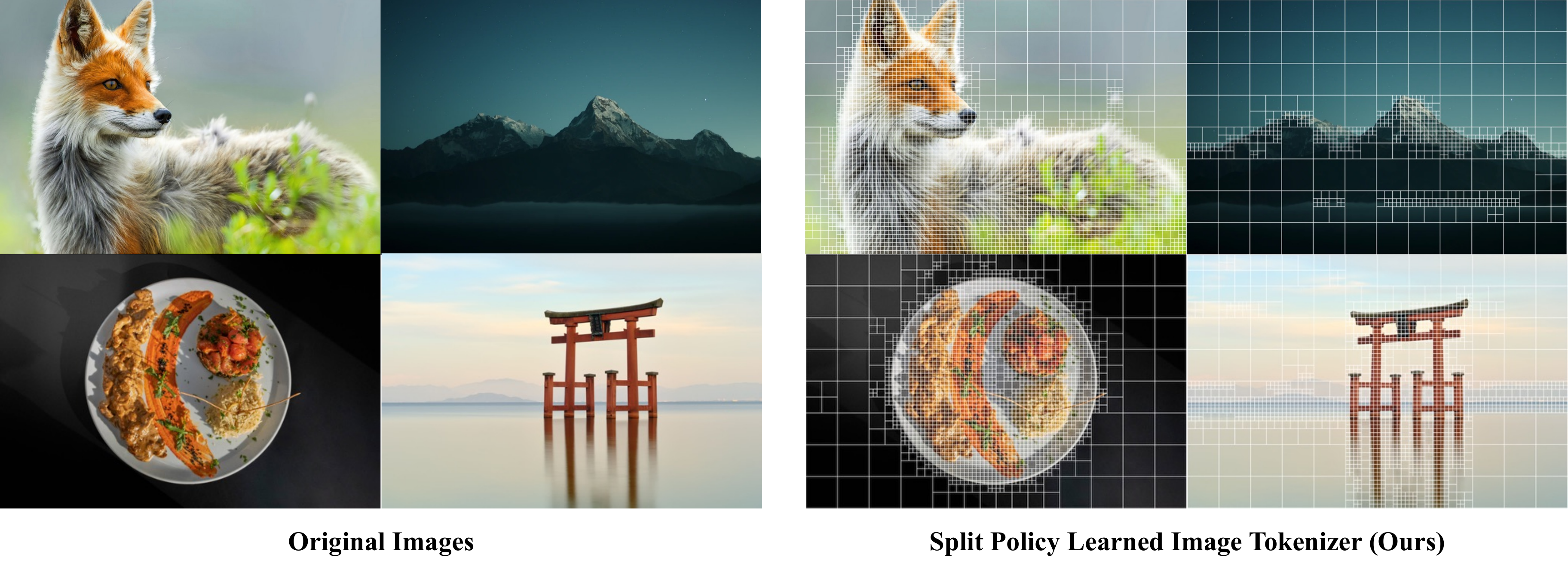}
\caption{\textbf{S}plit \textbf{P}olicy
\textbf{L}earned \textbf{I}mage \textbf{T}okenization.
SPLIT is an adaptive patchification framework that learns to allocate
fine tokens to semantically important regions and coarse tokens to
homogeneous ones, accelerating ViT inference without sacrificing
accuracy.}
\label{fig:teaser}
\end{center}
\vspace{-0.5cm}
\end{figure}

While promising, these methods are limited in their efficiency gains due to the cost of generating and processing dense tokens in the early
stages. An alternative approach is to perform adaptive
tokenization at the patch embedding stage itself. For example,
Quadformer~\citep{ronen2023vision} uses the feature discrepancy between an
image and its downsampled-then-upsampled counterpart as a proxy for
regional importance, and APT~\citep{choudhury2025apt} uses entropy as a heuristic to estimate task-relevant uncertainty, allocating higher
resolution to regions that are more informative for dense visual tasks.

Although adaptive patchification has proven effective, existing methods
still rely on hand-crafted or external proxies to estimate regional importance.
Visual signals such as entropy provide a useful estimate of local
uncertainty or information density, but they are not directly optimized
for the downstream task and may overlook semantically important regions
while mistakenly assigning high importance to visually complex but
task-irrelevant areas. The natural alternative, using semantic importance
derived from the ViT itself, is impractical, since it requires first
processing the full dense token sequence to obtain reliable importance
signals. This leads to a key question: \emph{can we obtain a lightweight
yet semantic importance signal before processing with ViT layers?}

Our key insight is that this dilemma is less rigid than it appears. Although
processing the full dense token sequence is expensive, coarse tokens
provide a lightweight alternative for obtaining model-aligned semantic
cues. By optimizing the patch selection policy through the downstream
task objective, the errors in the final prediction relative to different patch selection policies implicitly reveal which regions are semantically important, allowing the
splitting decisions themselves to be learned in a task-aware manner,
without requiring external proxies or costly dense ViT guidance.

Building on this insight, we propose \textbf{S}plit \textbf{P}olicy
\textbf{L}earned \textbf{I}mage \textbf{T}okenization (SPLIT), an adaptive
patchification framework that learns task-aware patch selection from
coarse to fine. As shown in \cref{fig:teaser}, SPLIT uses coarse tokens as lightweight semantic anchors
and selectively refines only the regions that are semantically important
to the downstream task. To further encourage efficiency, we formulate
patch refinement as a budget-constrained decision-making problem and
optimize the network with reinforcement learning, learning effective
region selection without exceeding the prescribed token budget.

Our contributions can be summarized as follows:
\begin{itemize}
    \item We introduce \textbf{SPLIT}, an adaptive patchification
    framework that learns task-aware patch selection from coarse semantic
    anchors, rather than pixel-level heuristics.

    \item We formulate adaptive patch refinement as a budget-constrained
    decision-making problem and optimize the decision policy with
    reinforcement learning, enabling effective region selection under a
    fixed token budget.

    \item SPLIT is lightweight to train: it adapts pretrained ViTs to
    semantically aligned tokenization with minimal fine-tuning, and
    transfers cleanly across image classification, object detection,
    semantic segmentation, and VQA benchmarks.
\end{itemize}

\section{Related Works}
\subsection{Efficient Vision Transformer}
Vision Transformers (ViTs)~\citep{dosovitskiy2021vit, touvron2021deit, liu2021swin} have become a standard
architecture for visual backbones across diverse domains, including image
classification~\citep{dosovitskiy2021vit}, object
detection~\citep{carion2020detr}, semantic
segmentation~\citep{strudel2021segmenter}, and vision-language
tasks~\citep{radford2021clip}. Despite their strong performance and
scalability, ViTs typically rely on a fixed patchification protocol: an input
image is uniformly divided into non-overlapping patches, each of which is
linearly projected into a visual token. This protocol makes the input sequence
length depend primarily on image resolution rather than image content,
assigning the same token budget to redundant background regions, repetitive
textures, and semantically important objects.
\vspace{-0.1in}
\subsection{ViT Tokens Reduction}
Due to the quadratic complexity of self-attention with respect to sequence
length~\citep{vaswani2017attention}, reducing the number of visual tokens has
become an important direction for improving ViT efficiency. Existing methods
can be roughly divided into two categories. The first category reduces tokens
after patchification, using intermediate model signals such as attention
scores~\citep{liang2022evit, yin2022adavit}, token similarity~\citep{bolya2023tome}, or
auxiliary importance predictors~\citep{rao2021dynamicvit, meng2022adavit, pan2021ia} as proxies for token
relevance. These methods prune less informative
tokens~\citep{rao2021dynamicvit, liang2022evit, yin2022adavit, kong2022spvit, tang2022patchslimming} or merge redundant token
representations~\citep{bolya2023tome, lee2024dtem, chen2023diffrate, wu2023ppt}, allowing later transformer layers to
operate on shorter sequences. While effective, these methods reduce tokens
only after dense processing and introduce dynamic indexing or gather-scatter
operations that can limit practical gains from optimized attention kernels
such as FlashAttention~\citep{dao2022flashattention}.
The other category reduces tokens before transformer processing by adapting
the patchification process itself. APT~\citep{choudhury2025apt} and related
adaptive patch sizing methods~\citep{havtorn2023msvit, ronen2023vision, chen2023cfvit}
assign larger patches to visually simple regions and smaller patches to
visually complex regions, producing shorter input token sequences without
mid-layer token manipulation. However, these methods typically rely on
hand-crafted pixel-level heuristics and task-specific threshold tuning, which
increases adaptation cost for each downstream setting.
\vspace{-0.1in}
\subsection{Semantic Token Patchification}
Besides requiring task-specific thresholds for patchification, current
adaptive patch sizing methods rely primarily on low-level pixel statistics to
determine patch granularity. For example, local entropy or visual complexity
is often used as a proxy for where fine-grained tokens should be allocated.
However, such signals do not necessarily reflect semantic importance \citep{hwang2025dynamic}. As
illustrated in \cref{fig:split_vs_apt_patch_viz}, APT-style methods can over-tokenize noisy backgrounds or
repetitive textures, while smooth but semantically meaningful regions can be
compressed too aggressively. To address this limitation, our method learns a
semantic patchification policy from the model's own feature representations.

\section{Method}

\begin{figure*}[t]
    \vspace{-0.1in}
    \centering
    \includegraphics[width=\linewidth]{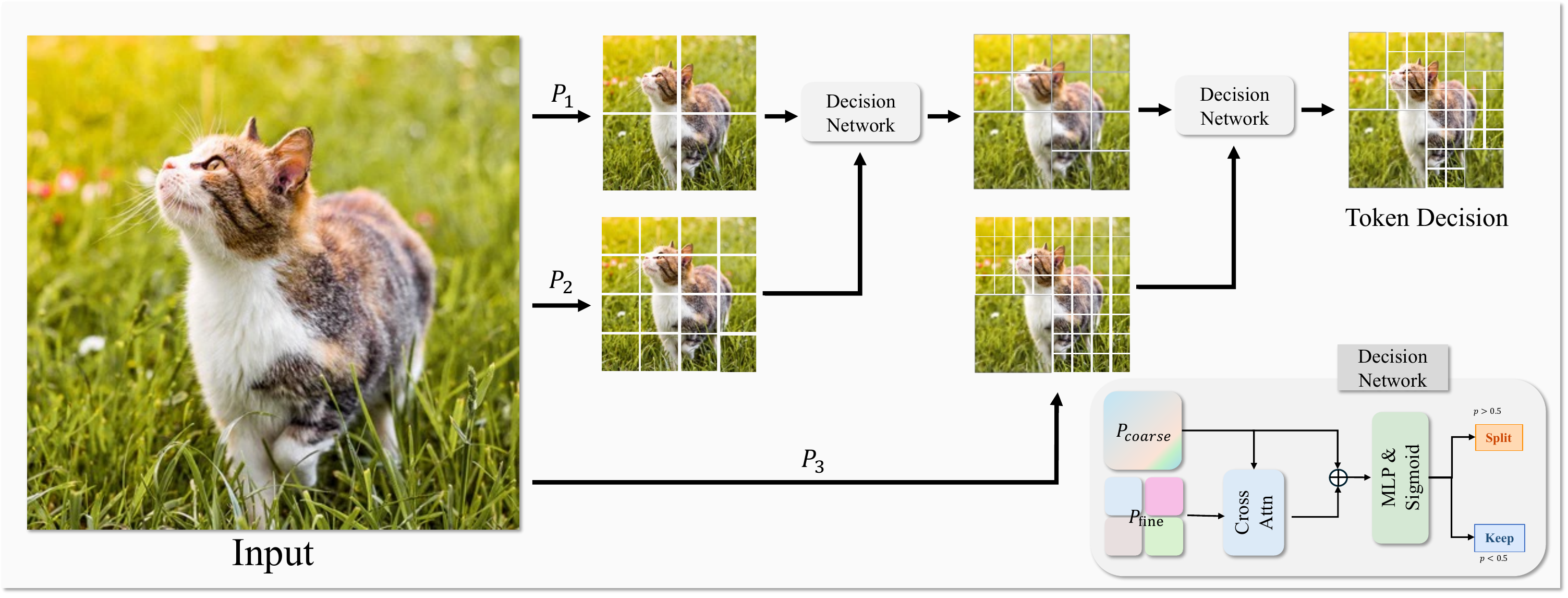}
    \caption{\textbf{Overview of SPLIT.} An input image is first divided into coarse patches. A decision network then performs $S$ refinement stages, deciding whether to split each parent patch into its children or keep it as a single token. The example uses two splitting stages, yielding a mixed-resolution token sequence. For each parent-child group, a lightweight cross-attention module pools the four child embeddings with the parent embedding as the query. The pooled summary is concatenated with the parent embedding and passed to a two-layer MLP to predict the split probability $p_i$.}
    \label{fig:pipeline}
\end{figure*}

SPLIT is a learnable adaptive patchification framework for efficient ViTs. Given an image $I \in \mathbb{R}^{3 \times H \times W}$, where $H$ and $W$ denote the image height and width, a vanilla ViT uniformly partitions it into $N = HW/p^2$ fixed-size patches, where $p$ is the patch size and $N$ is the resulting number of visual tokens. In contrast, SPLIT learns to generate only approximately $\rho N$ tokens under a token budget ratio $\rho \in (0,1]$.


\subsection{Problem Definition}
\label{ps}
Following prior work~\citep{ronen2023vision, choudhury2025apt}, we use a
multi-scale patch embedding function that maps patches of different
spatial sizes into a shared token embedding space, allowing adaptive
patches of varying resolutions to be processed by the same ViT backbone.

For a tokenizer with $S$ stages and smallest patch size $p$, we first
divide the image into patches of size $2^{S}p \times 2^{S}p$. A decision
network then determines, based on each patch's content, whether to split
it into four sub-patches of size $2^{S-1}p \times 2^{S-1}p$, and this
process is applied recursively to every split patch over $S$ rounds. The
result is a mixed-resolution tokenization that uses larger patches for
homogeneous regions and smaller patches for content-rich ones.

The decision network must trade off two conflicting objectives:
maximizing downstream performance (which favors more, finer tokens) and
minimizing compute per sample (which favors fewer, coarser tokens). We
formulate this as a constrained optimization problem and train the
decision network with reinforcement learning, minimizing the
unconstrained Lagrangian dual of the constraint.

\subsection{Decision Network}
\label{decision-module}
Consider a parent patch $P_i \in \mathbb{R}^{p \times p}$ with pre-positional
embedding $z_i^{p}\in\mathbb{R}^{d_{embed}}$, and let
$z_{i,1}^{c},\ldots,z_{i,4}^{c}\in\mathbb{R}^{d_{embed}}$ be the embeddings of
its four child sub-patches. All embeddings are computed using the multi-scale
patch aggregation scheme of APT~\citep{choudhury2025apt}.

The decision network is a lightweight cross-attention MLP that uses the parent
as a query and its children as keys and values, predicting the probability of
splitting the parent into its children:
\begin{align}
p_i = \sigma\!\left(h_\theta\big([\,z_i^{p};\;
\mathrm{CrossAttn}(z_i^{p}, \{z_{i,j}^{c}\}_{j=1}^{4})\,]\big)\right),
\end{align}
where $h_\theta$ is a two-layer MLP head and $\sigma$ is the sigmoid. The
attention pools the four child embeddings into a single summary, and the head
decides whether that summary differs sufficiently from the parent to justify
the finer tokens.

During training we sample the binary splitting decision
$s_i \sim \mathrm{Bernoulli}(p_i)$, where $s_i=0$ keeps the parent and $s_i=1$
replaces it with its four children. At inference we use the deterministic rule
$s_i=\mathbf{1}[p_i>0.5]$. For a single splitting stage this yields the
effective token set
\begin{align}
\mathcal{T}(x,s) =
\{z_i^{p} : s_i=0\}
\cup
\{z_{i,j}^{c} : s_i=1,\; j=1,\ldots,4\}.
\end{align}
With two stages, patches are organized into three scales (top, mid, fine), and
the same network is applied at both transitions to decide whether to split
each parent into its children. The formulation extends naturally to any number
of stages.

\subsection{PPO Training with Lagrange Dual}
\label{objective}

Training the decision network with ordinary supervised learning is not straightforward. In SPLIT, each coarse image patch must decide whether it should remain a single token or be split into finer subpatch tokens. This indexing operation is non-differentiable, so the decision network cannot be directly optimized through the downstream ViT loss by standard backpropagation. This non-differentiability can be addressed through differentiable relaxations, MSViT \citep{havtorn2023msvit} makes use of Straight-Through Estimators and Gumbel-Sigmoid, however such relaxations either need to materialize all coarse and fine tokens at training time or need to prune un-materialized tokens which results in the un-selected path not getting any gradients. Due to these limitations we approach this problem using policy gradients and avoid using biased differentiable relaxations. We provide more detail on this decision in Appendix section \ref{sec:supp_relaxations}. Following the RL formulation used in adaptive routing methods such as Ada-K~\citep{adak}, we formulate patch splitting as a one-step reinforcement learning problem.

Consider a single stage of splitting, let $x \in \mathcal{X}$ be an input image and let $\mathcal{P}(x)=\{p_i\}_{i=1}^{N}$ denote the set of coarse patches. The decision network $\pi_\theta$ outputs a Bernoulli splitting distribution for each coarse patch,
$\pi_\theta(m_i=1 \mid p_i), m_i \in \{0,1\},$, where $m_i=1$ means that patch $p_i$ is split into its child subpatches, and $m_i=0$ means that $p_i$ is retained as a coarse token. The full splitting mask is
$M = (m_1,\ldots,m_N) \sim \pi_\theta(\cdot \mid x)$. A sequence packing operation $\mathcal{T}(x,M)$ uses the mask to create the final token sequence combining the fine tokens with the un-split coarse tokens retained by $M$. $\mathcal{T}(x,M)$ is now the input to the vision transformer backbone $f_\phi$.

For a sampled mask $M$, we compute the downstream error signal
$\mathcal{E}(x,M)$ and the fraction of sequence length relative to using
all fine patches,
$\mathcal{P}(x,M) = |\mathcal{T}(x,M)| / |\mathcal{T}_{\mathrm{fine}}(x)|$,
where $\mathcal{T}_{\mathrm{fine}}(x)$ denotes the fully split patchify.
The decision network has two conflicting objectives: it should preserve
downstream task performance while using as few tokens as possible. We
write the constrained objective as
\begin{equation}
    \min_{\theta}\; J_{\mathrm{err}}(\theta)
    \quad \text{s.t.} \quad
    J_{\mathrm{tok}}(\theta) \leq \rho,
\end{equation}
where
$J_{\mathrm{err}}(\theta) =
\mathbb{E}_{x,M}[\mathcal{E}(x, M)]$ and
$J_{\mathrm{tok}}(\theta) =
\mathbb{E}_{x,M}[\mathcal{P}(x, M)]$ are taken over $x\sim\mathcal{D}$,
$M\sim\pi_\theta(\cdot\mid x)$, and $\rho$ is the target sequence length
fraction. This is analogous to the reward-cost formulation in Safe
RLHF~\citep{safe_rlhf}, where the main objective is optimized subject to
an expected cost constraint.

Using a Lagrangian relaxation, we optimize
\[
    \mathcal{L}(\theta,\lambda)
    =
    J_{\mathrm{err}}(\theta)
    +
    \lambda\big(J_{\mathrm{tok}}(\theta)-\rho\big),
    \qquad
    \lambda \geq 0.
\]
For policy optimization, the constant term $-\lambda\rho$ does not affect gradients, we also normalize the combined cost by $(1+\lambda)^{-1}$ to keep the scale of the cost stable as $\lambda$ changes.
\[
    C(x,M;\lambda)
    =
    \frac{\mathcal{E}(x, M)+\lambda \mathcal{P}(x, M)}{1+\lambda}.
\]
For each image $x$, we sample $L$ masks $M^{(1)},\ldots,M^{(L)} \sim \pi_{\theta_{\mathrm{old}}}(\cdot\mid x)$. We use a leave-one-out advantage comparing each of the $\ell$ rollouts to the other rollouts from the same image:
\[
    \widehat{A}_\ell
    =
    -(C_\ell - \frac{1}{L-1}\sum_{j \neq \ell} C_j)
    \qquad 
    \text{where,}
    \quad
    C_\ell
    =
    \frac{\mathcal{E}(x, M^{(\ell)})+\lambda \mathcal{P}(x, M^{(\ell)})}{1+\lambda}.
\] \vspace{-0.2in}

Thus, a rollout has a positive advantage if its combined cost is less than that of the other sampled masks. We then optimize the decision network with a PPO clipped objective. Let
$r_\ell(\theta)
    =
    \frac{\pi_\theta(M^{(\ell)}\mid x)}
    {\pi_{\theta_{\mathrm{old}}}(M^{(\ell)}\mid x)}$
be the importance ratio, the PPO loss is
\[
    \mathcal{L}_{\mathrm{PPO}}(\theta)
    =
    -
    \mathbb{E}_{x,\ell}
    \left[
    \min\left(
    r_\ell(\theta)\widehat{A}_\ell,
    \operatorname{clip}
    \big(r_\ell(\theta),1-\epsilon,1+\epsilon\big)
    \widehat{A}_\ell
    \right)
    \right].
\]
This objective allows the decision network to learn from discrete masks while preventing very large policy updates. Note that $\mathcal{L}_{\mathrm{PPO}}$ does not explicitly guide the decision network towards $\rho$, instead the multiplier $\lambda$ decides the relative importance of the two conflicting objectives and this decides the optimal $\mathbb{E}_{x\sim\mathcal{D},\,M\sim\pi_\theta(\cdot\mid x)} \big[\mathcal{P}(x, M))\big]$ corresponding to minimizing $\mathbb{E}_{x\sim\mathcal{D},\,M\sim\pi_\theta(\cdot\mid x)} \big[C(x, M; 
\lambda)]$

Effectively $\lambda$ is the only source of control on how aggressively our decision network chooses to split patches, if $\lambda$ is too small, the decision network may split nearly all patches and ignore the token budget. If $\lambda$ is too large, the network may retain too few fine tokens and collapse to an overly cheap representation. Therefore, following the dual descent update used in constrained RL~\citep{safe_rlhf}, we update $\lambda$ independently of the PPO step to ensure that $\lambda$ pushes the decision network to respect our budget $\rho$:
\vspace{-0.1in}
\[
    \lambda_{t+1}
    =
    \Big[
    \lambda_t
    +
    \eta_\lambda
    \big(
    \overline{P}_t-\rho
    \big)
    \Big]_+,
\quad \text{where } \quad
    \overline{P}_t
    =
    \frac{1}{BL}
    \sum_{b=1}^{B}\sum_{\ell=1}^{L}
    \mathcal{P}(x_b,M_b^{(\ell)}),
\] \vspace{-0.1in}
$B$ is the batch size, $\eta_\lambda$ is the dual learning rate, and $[\cdot]_+=\max(\cdot,0)$ enforces non-negativity. 

\subsection{Decision Network Warm-start}
\label{warmstart}

Arbitrary initialization of the decision network can make RL training
unstable. To reduce this instability, we warm-start the decision network
with deterministic pseudo-labels derived from patch-subpatch similarity.
Let $z_i^{c}$ be the embedding of a coarse patch $p_i$ and
$\{z_{i,k}^{f}\}_{k=1}^{4}$ be the embeddings of its four child
subpatches. For each parent--child pair, we compute the cosine
dissimilarity $d_{i,k} = (1 - \cos(z_i^{c}, z_{i,k}^{f}))/2 \in [0,1]$.
The pseudo-label
$\widetilde{m}_i = \mathbf{1}\!\left[\max_k d_{i,k} > \tau_{\mathrm{split}}\right]$
marks a parent for splitting if at least one of its children is
sufficiently dissimilar from it, where $\tau_{\mathrm{split}}$ is a
learnable threshold. The decision network is trained to imitate these
pseudo-labels with a binary cross-entropy loss, and the threshold
$\tau_{\mathrm{split}}$ is updated by dual ascent on the same budget gap
$\overline{P}_t - \rho$ used during PPO training, so that the warm-start
labels themselves track the target token budget.

During warm-start, these cosine-similarity masks are also used to fine-tune
the ViT backbone, allowing the backbone to adapt to the adaptive token
layout before the policy is optimized with PPO. In all subsequent decision
network training, the ViT backbone is frozen.

\section{Experiments}
\subsection{Image Classification}

\begin{wrapfigure}{r}{0.5\linewidth}
    \vspace{-0.3in}
    \centering
    \includegraphics[width=\linewidth]{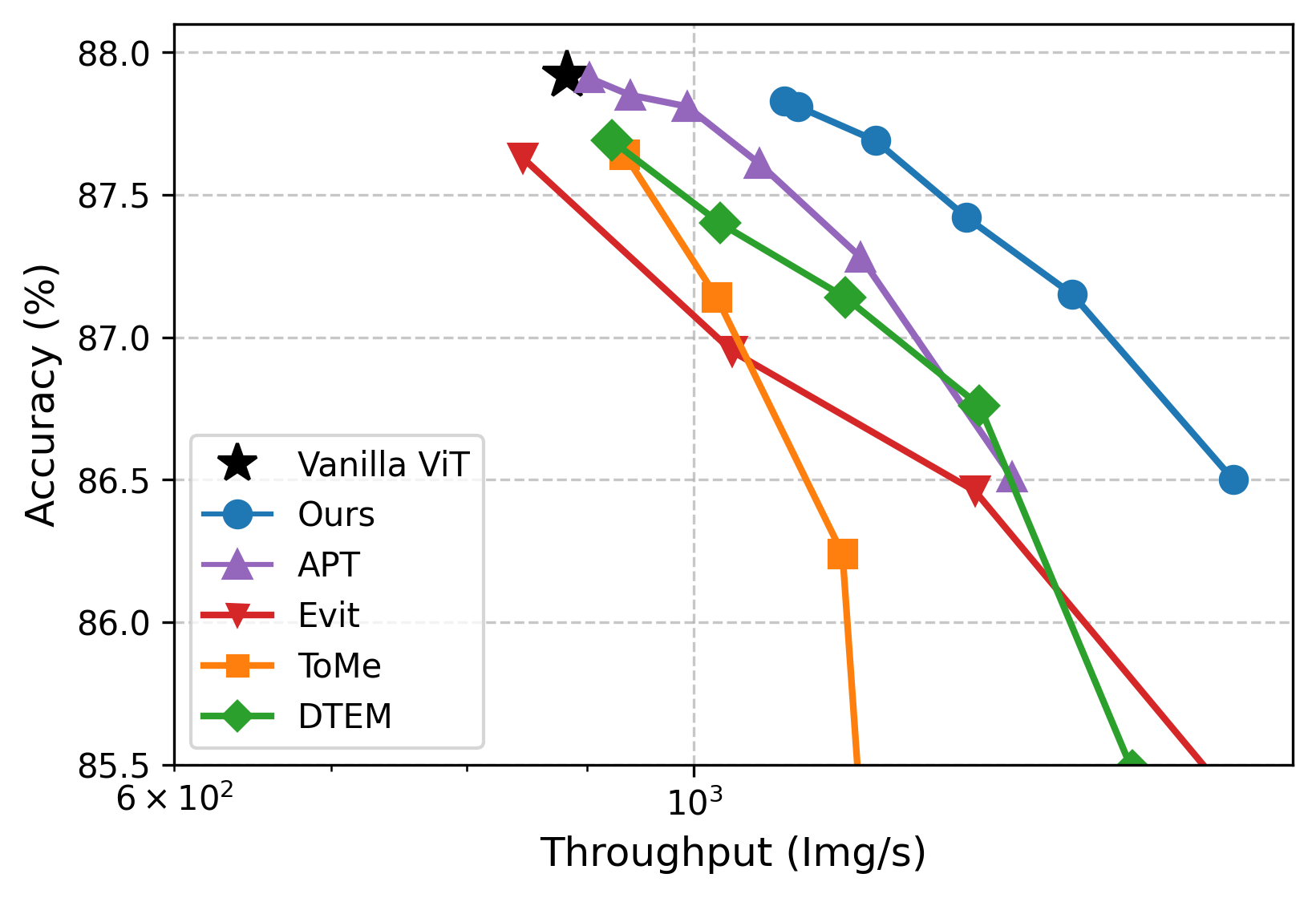}
    \caption{Comparison on ViT-L/14 at $224 \times 224$ resolution.}
    \label{fig:in1k-comparisons}
    \vspace{-0.1in}
\end{wrapfigure}

We provide results from training \footnote{We provide a detailed discussion on training cost in Appendix \cref{sec:supp_training_cost}} with SPLIT for 1 epoch of warm-start followed by 1 epoch of PPO training initialized from a checkpoint already finetuned on ImageNet \citep{deng2009imagenet}. We provide a comparison against baselines including token merging methods like EViT, ToMe and DTEM on ViT-L/14 at a resolution of 224 in \cref{fig:in1k-comparisons}. For a fair comparison in throughput, all the baselines in \cref{fig:in1k-comparisons} use flash attention. In \cref{tab:in1k-224,tab:in1k-336}, we show that our method is capable of matching performance to APT and vanilla ViT with significant speedup at an image size of 224 and 336.

\begin{table}[ht]
    \centering
    \begin{minipage}{0.49\linewidth}
    \centering
\begin{adjustbox}{width=\linewidth}
\begin{tabular}{lccccc}
    \toprule
    Model & Res/Patch & Img/s & Speedup & Acc & \\
    \midrule
    \textcolor{gray}{ViT-B} & \textcolor{gray}{224/16} & \textcolor{gray}{3310} & \textcolor{gray}{-} & \textcolor{gray}{85.1} & \\
    Random & 224/16 & 3751 & +13\% & 83.7 & \\
    APT-B & 224/16 & 3540 & +7\% & 85.1 & \\
    \rowcolor{blue!10}
    \textbf{SPLIT-B (Ours)} & 224/16 & 3635 & +10\% & 85.0 & \\
    \midrule
    \textcolor{gray}{ViT-L} & \textcolor{gray}{224/14} & \textcolor{gray}{883} & \textcolor{gray}{-} & \textcolor{gray}{87.9} & \\
    Random & 224/14 & 1049 & +19\% & 86.9 & \\
    APT-L & 224/14 & 993 & +12\% & 87.8 & \\
    \rowcolor{blue!10}
    \textbf{SPLIT-L (Ours)} & 224/14 & 1107 & +25\% & 87.8 & \\
    \midrule
    \textcolor{gray}{ViT-H} & \textcolor{gray}{224/14} & \textcolor{gray}{441} & \textcolor{gray}{-} & \textcolor{gray}{88.3} & \\
    Random & 224/14 & 568 & +29\% & 87.4 & \\
    APT-H & 224/14 & 542 & +23\% & 88.3 & \\
    \rowcolor{blue!10}
    \textbf{SPLIT-H (Ours)} & 224/14 & 533 & +21\% & 88.2 & \\
    \bottomrule
\end{tabular}
\end{adjustbox}
\vspace{0.1in}
\captionof{table}{\textbf{ImageNet Classification at 224.} Comparison of ViT models with different patch reduction strategies at 224x224 resolution.}

\vspace{-0.1in}
\label{tab:in1k-224}
    
    \end{minipage} %
    \hfill
    \begin{minipage}{0.49\linewidth}
    \centering
\begin{adjustbox}{width=\linewidth}
\begin{tabular}{lccccc}
    \toprule
    Model & Res/Patch & Img/s & Speedup & Acc & \\
    \midrule
    \textcolor{gray}{ViT-L} & \textcolor{gray}{336/14} & \textcolor{gray}{395} & \textcolor{gray}{-} & \textcolor{gray}{88.2} & \\
    Random & 336/14 & 550 & +39\% & 87.3 & \\
    APT-L & 336/14 & 527 & +33\% & 88.1 & \\
    \rowcolor{blue!10}
    \textbf{SPLIT-L (Ours)} & 336/14 & 570 & +44\% & 88.1 & \\
    \midrule
    \textcolor{gray}{ViT-H} & \textcolor{gray}{336/14} & \textcolor{gray}{175} & \textcolor{gray}{-} & \textcolor{gray}{88.5} & \\
    Random & 336/14 & 272 & +55\% & 87.0 & \\
    APT-H & 336/14 & 263 & +50\% & 88.4 & \\
    \rowcolor{blue!10}
    \textbf{SPLIT-H (Ours)} & 336/14 & 281 & +61\% & 88.4 & \\
    \bottomrule
\end{tabular}
\end{adjustbox}
\vspace{0.1in}
\captionof{table}{\textbf{ImageNet Classification at 336.} Comparison of ViT models with different patch reduction strategies at 336x336 resolution.}
\vspace{-0.1in}
\label{tab:in1k-336}
    
    \end{minipage}
\end{table}


\begin{table}[t]
    \centering
    \begin{minipage}{0.49\linewidth}
    
\centering
\begin{adjustbox}{width=\linewidth}
\begin{tabular}{lccccc}
    \toprule
    Model & Res & Img/s & Speedup & mAP & AP50 \\
    \midrule
    \textcolor{gray}{EVA-02-B} & \textcolor{gray}{1536} & \textcolor{gray}{3.66} & \textcolor{gray}{-}& \textcolor{gray}{58.93} & \textcolor{gray}{77.85} \\
    Resizing & 1536 & 4.14  & +13\% & 58.43 & 77.22 \\
    APT & 1536 & 4.14 & +13\% & 58.79 & 77.65 \\
    \rowcolor{blue!10}
    \textbf{SPLIT (Ours)} & 1536 & 4.23 & +16\% & 58.84 & 77.80 \\
    \midrule
    \textcolor{gray}{EVA-02-L} & \textcolor{gray}{1536} & \textcolor{gray}{1.61}  &\textcolor{gray}{-}& \textcolor{gray}{62.28} & \textcolor{gray}{80.80} \\
    Resizing & 1536 & 1.91 & +19\% & 61.75 & 80.27 \\
    APT & 1536 & 1.91 & +19\% & 62.07 & 80.64 \\
    \rowcolor{blue!10}
    \textbf{SPLIT (Ours)} & 1536 & 2.02 & +25\% & 62.24 & 80.78 \\
    \bottomrule
\end{tabular}
\end{adjustbox}
\vspace{0.1in}
\captionof{table}{\textbf{Transfer to Object Detection.} SPLIT can be scaled to high-resolution dense image tasks supporting window attention.}
\label{tab:detection}
    
    \end{minipage} %
    \hfill
    \begin{minipage}{0.49\linewidth}
    \centering
\begin{adjustbox}{width=\linewidth}
\begin{tabular}{lccccc}
    \toprule
    Model & Res & Img/s  &Speedup&  aAcc & mIoU \\
    \midrule
    \textcolor{gray}{EVA-02-L} & \textcolor{gray}{512} & \textcolor{gray}{6.48}  &\textcolor{gray}{-}& \textcolor{gray}{86.67} & \textcolor{gray}{59.77} \\
    Resizing & 512 & 7.19 & +11\%& 86.09 & 58.81 \\
    APT & 512 & 7.19  &+11\%& 86.68 & 59.70 \\
    \rowcolor{blue!10}
    \textbf{SPLIT (Ours)} & 512 & 7.43 & +15\% & 86.53 & 59.65 \\
    \midrule
    \textcolor{gray}{EVA-02-L} & \textcolor{gray}{640} & \textcolor{gray}{3.44}  &\textcolor{gray}{-}& \textcolor{gray}{86.83} & \textcolor{gray}{60.05} \\
    Resizing & 640 & 3.82  &+11\%& 86.06 & 58.83 \\
    APT & 640 & 3.82  &+11\%& 86.82 & 60.01 \\
    \rowcolor{blue!10}
    \textbf{SPLIT (Ours)} & 640 & 3.97 & +15\% & 86.75 & 59.99 \\
    \bottomrule
\end{tabular}
\end{adjustbox}
\vspace{0.1in}

\captionof{table}{\textbf{Transfer to Semantic Segmentation.} SPLIT can handle pixel-level fine-grained tasks without compromising visual acuity.}
\label{tab:segmentation}
    
    \end{minipage}
\end{table}

\vspace{-0.2in}
\subsection{Dense Visual Tasks}

\paragraph{Object detection.}
Object detection is a strong stress test for adaptive patchification because localization depends on preserving fine spatial detail at high resolution.
To evaluate this setting, we train EVA-02 backbones~\citep{fang2024eva} with a ViTDet-style detection head~\citep{li2022vitdet} under window attention, and benchmark on COCO~\citep{lin2014microsoft} at $1536\times1536$.
As shown in \cref{tab:detection}, SPLIT stays in the same high-accuracy regime as the fixed-grid baseline and APT while delivering the best throughput across both model scales.
Notably, this throughput gain is achieved even when SPLIT and APT operate at a similar token-reduction level (about $30\%$): our runtime breakdown confirms that the remaining gap comes from lower decision making overhead in SPLIT, whereas APT incurs higher cost from entropy-based multi-scale scoring when extended beyond two patch sizes.
These results indicate that input-level learned policy remains effective for high-resolution dense detection with window attention.

\paragraph{Semantic segmentation.}
We further evaluate SPLIT on semantic segmentation, which requires fine-grained boundary understanding in addition to region-level consistency.
Following the EVA-02 segmentation protocol~\citep{fang2024eva}, we use an UperNet decoder~\citep{xiao2018upernet} and report results on ADE20K~\citep{zhou2019ade20k,zhou2017ade20k}.
\cref{tab:segmentation} shows that SPLIT closely matches baseline/APT quality on both aAcc and mIoU while consistently improving throughput.
In this task, the speed gain aligns with stronger sequence reduction: SPLIT uses roughly $35\%$ fewer tokens, compared with about $30\%$ for APT, yielding additional runtime savings without materially affecting segmentation quality.
Together, these results show that SPLIT preserves pixel-level performance while improving efficiency.

\begin{table*}[t]
    \centering
    \vspace{-0.1in}
    \begin{adjustbox}{width=\linewidth}
    \begin{tabular}{lccccccccccc}
        \toprule
        Model & Img/s & Speedup & $\text{VQA}^{v_2}$ & GQA & $\text{SQA}^{I}$ & $\text{VQA}^{T}$ & POPE & MME & MMB & $\text{MMB}^{C}$ & MMV\\
        \midrule
        \textcolor{gray}{LLaVA-1.5-7B} & \textcolor{gray}{3.70} & \textcolor{gray}{-} & \textcolor{gray}{78.5} & \textcolor{gray}{61.0} & \textcolor{gray}{67.8} & \textcolor{gray}{58.2} & \textcolor{gray}{86.9} & \textcolor{gray}{1510.1} & \textcolor{gray}{64.6} & \textcolor{gray}{58.1} & \textcolor{gray}{30.7} \\
        Resizing & 4.51 & +22\% & 77.5 & 61.1 & 66.8 & 56.5 & 86.6 & 1473.8 & 63.2 & 58.1 & 30.2 \\
        APT & 4.51 & +22\% & 77.9 & 61.4 & 67.5 & 56.9 & 86.4 & 1474.0 & 63.8 & 58.2 & 30.8 \\
        \rowcolor{blue!10}
        \textbf{SPLIT (Ours)} & 4.37 & +18\% & 78.6 & 61.7 & 69.0 & 58.1 & 86.6 & 1497.3 & 65.2 & 58.9 & 31.3 \\
        \midrule
        \textcolor{gray}{LLaVA-1.5-13B} & \textcolor{gray}{2.22} & \textcolor{gray}{-} & \textcolor{gray}{80.0} & \textcolor{gray}{63.2} & \textcolor{gray}{72.7} & \textcolor{gray}{61.2} & \textcolor{gray}{87.1} & \textcolor{gray}{1530.6} & \textcolor{gray}{68.5} & \textcolor{gray}{63.4} & \textcolor{gray}{35.4} \\
        Resizing & 2.72 & +23\% & 78.9 & 61.1 & 72.0 & 59.1 & 86.8 & 1496.9 & 65.8 & 62.5 & 33.9 \\
        APT & 2.72 & +23\% & 79.4 & 63.0 & 72.4 & 59.5 & 87.2 & 1511.2 & 66.5 & 63.7 & 34.7 \\
        \rowcolor{blue!10}
        \textbf{SPLIT (Ours)} & 2.66 & +20\% & 79.6 & 63.1 & 73.4 & 60.2 & 86.9 & 1504.9 & 67.5 & 64.0 & 36.3 \\
        \bottomrule
    \end{tabular}
    \end{adjustbox}
    \captionof{table}{\textbf{Transfer to VQA}. SPLIT enables significant throughput increase while matching or exceeding performance to the baseline.}
    \vspace{-0.2in}
    \label{tab:llava}
\end{table*}

\paragraph{Visual question answering.}
We further evaluate the transferability of SPLIT to multimodal visual
question answering by applying it to the vision backbone of
LLaVA-1.5~\citep{liu2024llava15}. Following APT~\citep{choudhury2025apt}, we
replace the original uniform patch embedding with our learnable adaptive
patchification module, keeping the language model frozen while training the
decision module and the vision-language projection layer. As shown in
\cref{tab:llava}, SPLIT consistently improves inference throughput over
vanilla LLaVA while matching or exceeding the baseline accuracy across a
wide range of VQA benchmarks \citep{goyal2017vqav2, hudson2019gqa, lu2022scienceqa, singh2019textvqa, schwenk2022pope, fu2024mme, liu2024mmbench, yu2023mmvet}. Although APT achieves a slightly higher
speedup at the same target token fraction, it falls short of the baseline on
most metrics, whereas SPLIT recovers or surpasses the baseline on nearly
every benchmark for both the 7B and 13B variants.


\subsection{Ablation}
\label{sec:ablation}

\begin{wrapfigure}{r}{0.6\linewidth}
    \centering
    \includegraphics[width=1.0\linewidth]{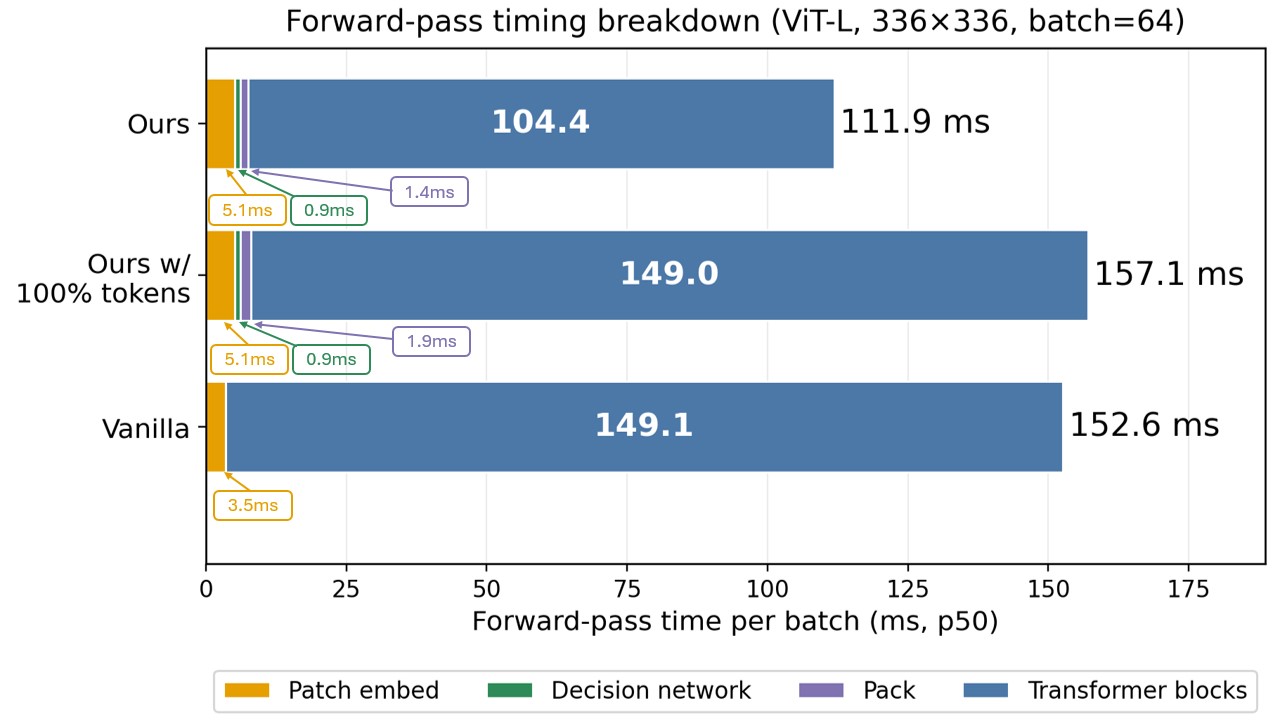}
    \caption{\textbf{Forward-pass Timing Breakdown.} SPLIT incurs minimal overhead from decision network and patch splitting operations while saving on transformer blocks, thus resulting in a significant throughput increase}
    \label{fig:inference_breakdown}
    \vspace{-0.2in}
\end{wrapfigure}

\paragraph{Measuring SPLIT overhead.}
We next measure the computational overhead introduced by SPLIT. Adaptive
patchification breaks the constant-input-shape assumption that GPUs are
heavily optimized for, and the decision network itself adds a small
forward pass on top of patch embedding. Both costs need to be accounted
for before any throughput claims can be trusted. We report this analysis
in \cref{fig:inference_breakdown}. With no sequence reduction applied, SPLIT incurs a minor 3\% overhead relative to the vanilla ViT baseline due to multi-scale patch embedding computations, decision network and additional packing logic to join the multi-scale patch embeddings. Once
sequence reduction is enabled, the savings from processing fewer tokens
substantially outweigh the overhead.

\begin{figure}[ht]
\centering

\begin{minipage}{0.49\linewidth}
    \centering
    \includegraphics[width=\linewidth]{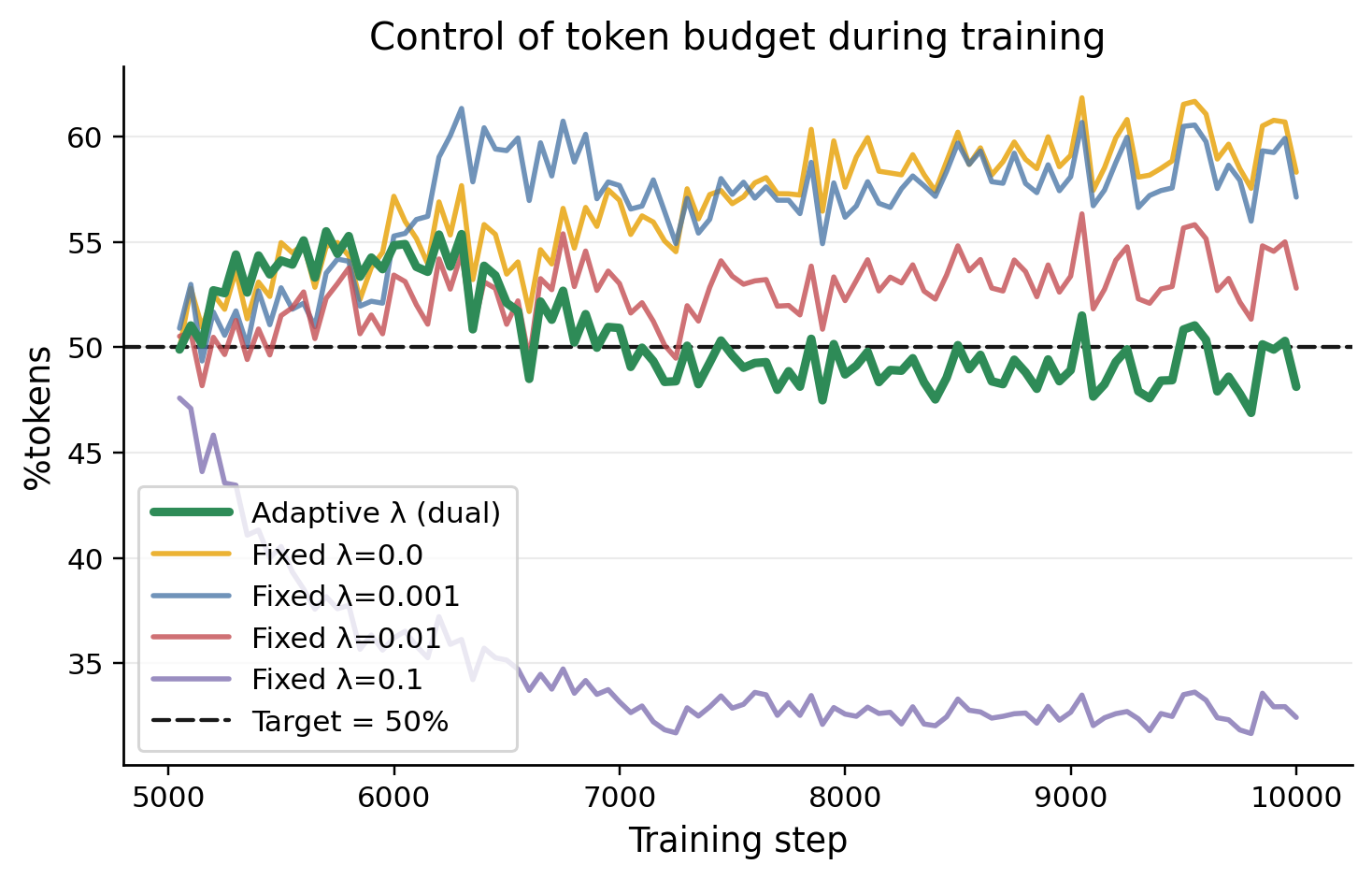}
    \caption*{(a) Token-budget dynamics during training.}
\end{minipage}
\hfill
\begin{minipage}{0.49\linewidth}
    \centering
    \includegraphics[width=\linewidth]{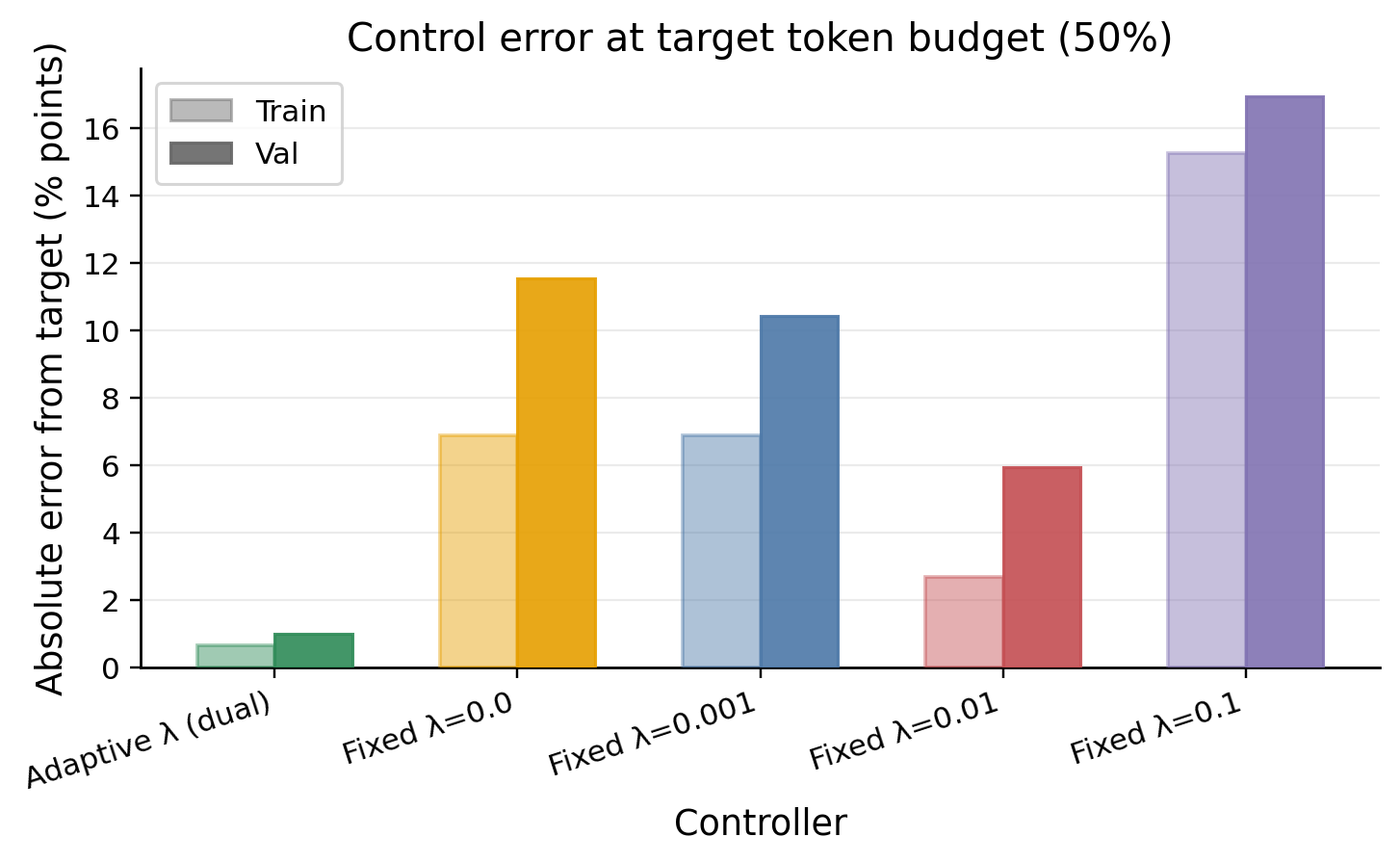}
    \caption*{(b) Train and validation control error.}
\end{minipage}

\caption{
Adaptive $\lambda$ more reliably controls the token budget than fixed-$\lambda$ baselines.
Left: percent tokens used during training, with the dashed line indicating the 50\% target.
Right: absolute deviation from the target token budget on train and validation sets.
}
\label{fig:lambda_control_ablation}
\vspace{-0.2in}
\end{figure}

\paragraph{Lagrange Multiplier Control.} As discussed in Section \ref{objective}, the multiplier $\lambda$ decides the relative weighting of the performance cost and compute cost. In figure \ref{fig:lambda_control_ablation} we show that a continuously updated $\lambda$ actively guides the model towards the specified token budget. Using a fixed value of $\lambda$ for training involves laborious tuning of the value of $\lambda$ to achieve the desired token budget. We also observe that even though we can achieve similar \% tokens using fixed $\lambda=0.01$ and adaptive $\lambda$, the gap in control error between train and val sets is much smaller when using an adaptive $\lambda$ indicating that it is a better control mechanic.

\begin{figure}[ht]
    \begin{subfigure}[t]{0.49\linewidth}
        \centering
        \includegraphics[width=\linewidth]{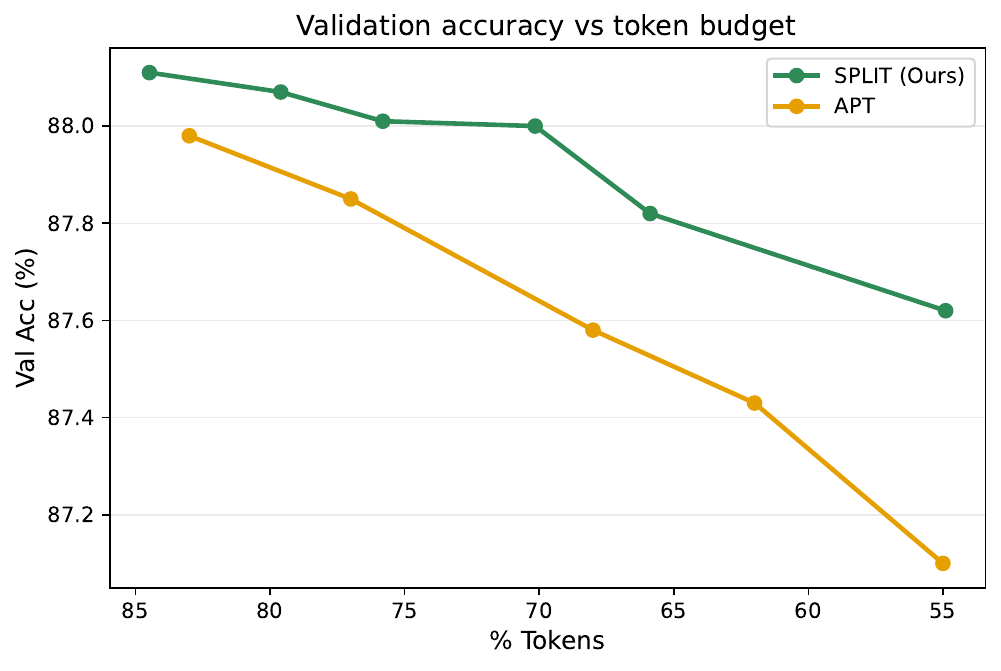}
        \caption{Accuracy vs.\ \% tokens for ViT-L 336/14}
        \label{fig:apt_pareto}
    \end{subfigure}
    \label{fig:apt_vs_split}
    \hfill
    \centering
    \begin{subfigure}[t]{0.49\linewidth}
        \centering
        \includegraphics[width=\linewidth]{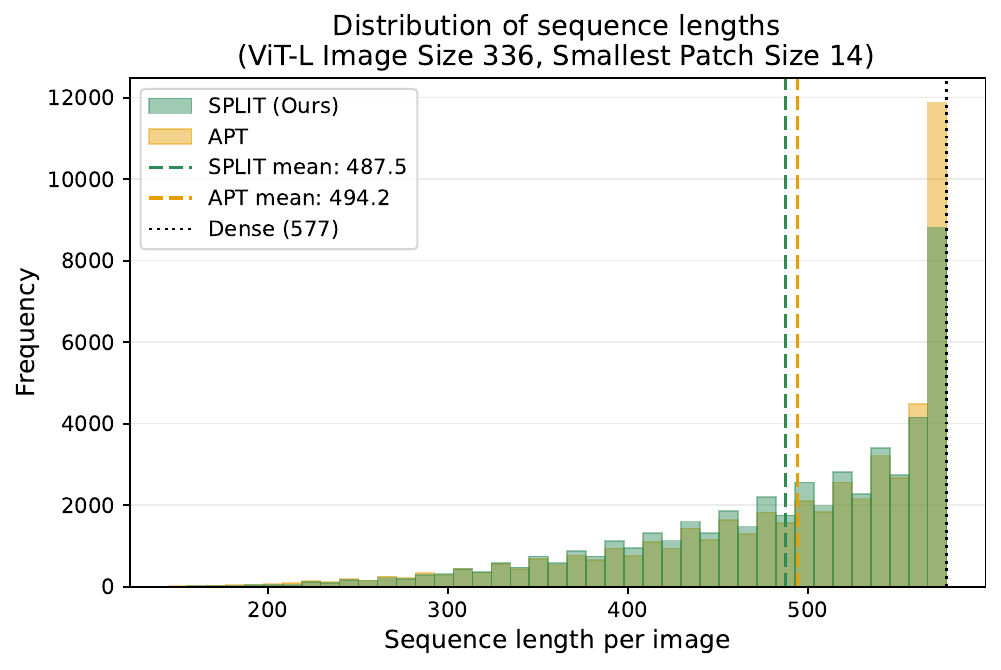}
        \caption{Sequence length distribution at same mean \% tokens}
        \label{fig:apt_seq_length_dist}
    \end{subfigure}
    \caption{Comparison of our method against APT.}
    \vspace{-0.1in}
\end{figure}

\begin{figure}[t]
    \centering
    \includegraphics[width=\linewidth]{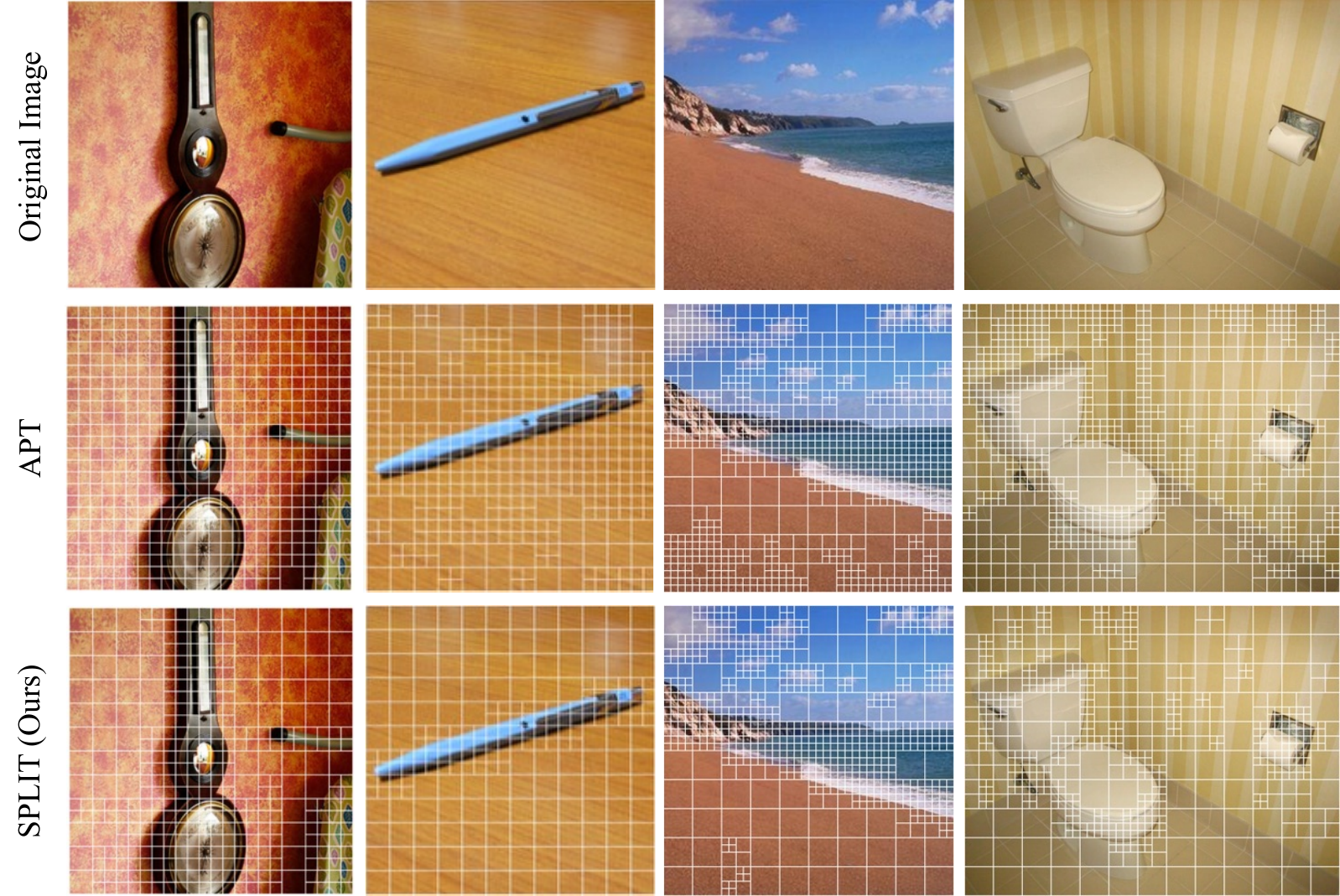}
    \caption{\textbf{Patch allocation comparison.} APT (middle) over-tokenizes
    high-entropy textures regardless of their semantic content, while SPLIT (bottom) concentrates fine tokens around objects.}
    \label{fig:split_vs_apt_patch_viz}
    \vspace{-0.25in}
\end{figure}

\subsection{Comparison against APT}
We compare SPLIT directly against APT under matched conditions to isolate
the contribution of our decision network. Since both methods use the same
multi-scale patch embedding scheme, any performance gap must come from
where each method chooses to allocate fine tokens. \cref{fig:apt_pareto}
shows the Pareto frontier of accuracy against token retention on ViT-L at
$336\times336$. Across every token budget we evaluate, SPLIT
achieves higher accuracy than APT, with the gap widening as the token
budget tightens.

\cref{fig:apt_seq_length_dist} examines this allocation behavior more closely
by plotting the per-image sequence length distributions of the two
methods, with their mean sequence lengths matched. Both distributions are
skewed toward maximum-length sequences, indicating that a substantial
fraction of images are split fully, but APT does this noticeably more
often than SPLIT. The mass that SPLIT pulls away from full splits is
redistributed toward lower sequence lengths, meaning SPLIT identifies more
images for which fewer tokens suffice. APT's tendency to over-tokenize is
consistent with its reliance on pixel entropy: visually busy images
trigger fine splits regardless of whether the busy regions are
semantically informative.

\cref{fig:split_vs_apt_patch_viz} illustrates this difference qualitatively.
SPLIT concentrates fine tokens around object boundaries and
semantically meaningful structure, while leaving textured but
task-irrelevant regions such as patterned walls, tiled floors and repeated
backgrounds, at coarse resolution. In contrast, APT frequently
allocates fine tokens to these high-entropy but semantically empty
regions, consuming budget that SPLIT redirects to objects. 
Together,
these results support our central claim: pixel-level heuristics
mis-locate tokens in predictable ways, and a learned decision policy
trained against the downstream objective improves patch size allocations.

\paragraph{Quantifying semantic alignment.}
To make this comparison concrete, we measure \emph{fine-token recall} on ADE20K. We define object recall as the fraction of object-class patches that receive a fine ($16\times16$) token, and boundary recall as the fraction of class-boundary patches that receive a fine token, both at a matched token budget (full definition in Appendix \cref{sec:supp_semantic}). As \cref{tab:semantic_recall} shows, SPLIT routes fine tokens to a much larger share of task-relevant patches than the entropy heuristic.

\begin{table}[ht]
\centering
\caption{\textbf{Fine-token recall on ADE20K} at a matched token budget
($\sim\!0.7$). Object and boundary recall measure the fraction of object-class and
class-boundary patches, respectively, that receive a fine token.}
\label{tab:semantic_recall}
\begin{tabular}{lccc}
\toprule
\textbf{Method} & \textbf{\% Tokens} & \textbf{Object recall} & \textbf{Boundary recall} \\
\midrule
Random        & 0.700 & 0.697 & 0.698 \\
APT (entropy) & 0.719 & 0.808 & 0.833 \\
\rowcolor{blue!10}
\textbf{SPLIT (Ours)} & 0.714 & \textbf{0.907} & \textbf{0.945} \\
\bottomrule
\end{tabular}
\end{table}

\section{Conclusion}
\vspace{-0.05in}
\label{sec:conclusion}
We presented \textbf{SPLIT}, an adaptive patchification framework that learns
a content-based patch splitting policy. By treating coarse tokens as lightweight semantic
anchors and selectively refining only the regions that matter for the
downstream task, SPLIT produces shorter input sequences that are better
matched to task-relevant content. We formulate patch refinement as a
budget-constrained decision problem and train the splitting policy with
reinforcement learning, allowing the model to learn where finer tokens are
most useful under a fixed sequence-length budget.

Across ImageNet classification, COCO detection, and ADE20K segmentation, SPLIT
matches or surpasses the accuracy of full-resolution ViT baselines while
delivering consistent throughput gains over prior adaptive patchification
methods. These results show that grounding patch splitting decisions in the model's interpretation of task relevant information
generalises well
beyond classification and offers a practical path toward more efficient ViTs
without sacrificing task performance.

\noindent\textbf{Limitations.} SPLIT delivers significant inference speedups by reducing the average token sequence length to a target reduction ratio that is fixed at training time. Evaluating SPLIT for a different desired speedup requires a short retraining run rather than a free inference-time hyperparameter. Conditioning the patchification policy on the desired ratio at inference would remove this constraint. Our second limitation is scope: we evaluate SPLIT only on image understanding tasks. Image generation typically operates at much higher resolutions and with much larger models, a regime where our method's speedup is most pronounced. This remains untested, and we view extending SPLIT to generative settings as a promising direction for future work.

\noindent\textbf{Future Work.} Beyond the directions discussed in the limitations, using SPLIT for self-supervised tasks to create policies reusable across multiple tasks is a promising direction. On a similar note, a thorough analysis on the behavior of SPLIT trained on a task and re-used on another is yet to be characterized and may be useful in understanding how to develop a task-agnostic tokenization policy.

\begin{ack}
This work was supported by Fujitsu Research of America. We thank Souraja Kundu for helpful feedback and assistance with the writing of this paper.
\end{ack}

\bibliographystyle{plainnat}
\bibliography{reference}
\newpage
\appendix
\section{Implementation Details}
\label{sec:supp_impldetails}

\paragraph{Hardware Setup.}
All inference-time results, including throughput analysis, were
computed on a single NVIDIA A100. ImageNet classification training was
conducted on a single node of 8x NVIDIA A100s. Object detection, semantic
segmentation, and visual QA experiments were conducted on a single node of 8x
NVIDIA H200s. We used a single node for all work in this paper.

\paragraph{Layer-level Merging Baselines.}
For comparisons against layer-level token reduction methods, we used the
official repositories for EViT~\citep{liang2022evit},
ToMe~\citep{bolya2023tome}, and DTEM~\citep{lee2024dtem}. We kept all
experimental settings (training schedule, optimizers, input resolutions, and
other hyperparameters) identical to the original baselines to ensure a fair
comparison.

\paragraph{Image Classification.}
We implemented image classification models using the \texttt{timm}
library~\citep{timm}, leveraging its pretrained checkpoints. The ImageNet-1K
dataset~\citep{deng2009imagenet} was used for training and evaluation,
following prior work~\citep{havtorn2023msvit, ronen2023vision}. We perform 1 epoch of warmstart and 1 epoch of RL fine-tuning for all experiments. No augmentations
are used. All training
was conducted on 8 GPUs with a batch size of 512. For the decision network,
we used a warm-start learning rate of $1\mathrm{e}{-3}$, a PPO learning
rate of $1\mathrm{e}{-5}$, a Lagrangian dual learning rate of
$5\mathrm{e}{-4}$ and a warm start threshold update rate of $\eta_\tau = 1\mathrm{e}{-3}$. All image classification experiments involve a single stage of splitting.

\paragraph{Visual QA.}
For our Visual Question Answering (VQA) experiments, we utilized the official
LLaVA-1.5~\citep{liu2024llava15} implementation and its pretrained checkpoints.
Unlike the original approach, which fine-tunes on the entire dataset for one
epoch, we fine-tuned only 5\% of the dataset and do one epoch for warmstart and one epoch for RL. To adapt to this setting, we reduced the learning rate
by a factor of 10 while following all other fine-tuning procedures recommended by LLaVA. The base image resolution was set to 336 with a patch size of 14, as specified in LLaVA's default configuration. We keep our retained token fraction to $0.7$ and $0.001$ for warm start threshold update rate.

\paragraph{Object Detection.}
For object detection, we used the official implementation of
EVA-02~\citep{fang2024eva} along with its pretrained checkpoints, which
utilize a window attention mechanism. Fine-tuning was conducted following
the recommended procedures outlined in EVA-02. Consistent with our
previous experiments, we fine-tuned for 5\% of the total iterations,
split evenly between the warm-start and PPO phases, while reducing the
learning rate by a factor of 10. Following EVA-02's settings, the image
resolution was 1536 with a base patch size of 16. SPLIT additionally
produces patches at sizes 128, 64, 32, and 16. For the decision network,
we used a warm-start learning rate of $1\mathrm{e}{-3}$, a PPO learning
rate of $1\mathrm{e}{-5}$, and a Lagrangian dual learning rate of
$5\mathrm{e}{-3}$, with a target retained token fraction of $\rho = 0.7$
and a warm start threshold update rate of $\eta_\tau = 1\mathrm{e}{-3}$.

\paragraph{Semantic Segmentation.}
We also utilized the official EVA-02 implementation along with its
pretrained checkpoints for semantic segmentation. The ADE20K
dataset~\citep{zhou2019ade20k, zhou2017ade20k} was used for training and
evaluation. Fine-tuning followed the recommended procedures outlined in
EVA-02. In alignment with our previous experiments, we fine-tuned for
5\% of the total iterations, split evenly between the warm-start and PPO
phases, while reducing the learning rate by a factor of 10. According to
EVA-02's settings, the image resolution was either 512 or 640, with a
base patch size of 16. SPLIT produces patches at sizes 64, 32, and 16.
The decision network hyperparameters match the detection setting, except
that the target retained token fraction is set to $\rho = 0.65$.

\section{Choice of Policy Gradients over Differentiable Relaxations}
\label{sec:supp_relaxations}

In this section, we provide more detail on our decision to use policy gradients which shape our core algorithm. As discussed in Section \ref{objective}, training the decision network requires gradients to flow through an indexing operation that selects whether a patch is represented by a coarse token or by 4 fine tokens in the final sequence passed to the transformer backbone. This operation is non-differentiable as the decision network makes a binary decision and that decision determines which tokens exist in the sequence. Using a differentiable relaxation possible in principle, MSViT \citep{havtorn2023msvit} builds one for mixed-scale tokenization. However, since the difficulty is that our discrete decision is structural, a relaxation then has only two options: 

\paragraph{Option 1: Keep both scales.} A soft mixture, or a masked union as in DynamicViT \citep{rao2021dynamicvit}, lets gradients flow, but every scale must be materialized. For a single splitting stage, this means materializing $5N$ tokens ($N$ coarse $+~4N$ fine), versus the $4N$ of the dense sequence. With attention cost quadratic in sequence length and activation memory growing with it, this becomes prohibitive at $1536\times1536$ resolution. Hierarchical splitting makes this worse: a fine token exists only if every ancestor split, so its relaxed weight is a product of gates across levels and gradients to each gate are filtered through the approximations of the others.

\paragraph{Option 2: Compute only the selected scale.} Drop the un-selected tokens to keep the training costs from increasing beyond the dense sequence limit and use Straight-Through Estimators for the binary decision. In this case the gradients are biased and the unselected tokens produce no learning signal and the policy can never directly learn whether a coarse token would have sufficed and must rely on stochastic exploration for this.

A differentiable relaxation produces a biased gradient estimator and must choose between materializing sequences longer than dense sequences or losing signal on the unselected branch of the decision. Policy gradients avoid both problems and trains on exactly the hard, reduced sequences used at inference, with unbiased gradients. Ada-K routing \citep{adak} chose PPO for the same structural problem in MoE routing, stating "direct optimization … is precluded due to the non-differentiable nature of the sampling operation". Also, methods where differentiable gating was tried needed extra stabilizers: MSViT's batch-shaping loss against gate collapse, DynamicViT's distillation loss.

\paragraph{What the RL formulation contributes.} The RL phase of training using policy gradients on hard sequences  yields measurable gains. Table \ref{tab:iter_det} shows the PPO phase improves object detection mAP monotonically over a warm-start-only policy ($58.06 \rightarrow 58.84$ at a matched budget). Table \ref{tab:semantic_recall_breakdown} shows that it sharpens semantic allocation by $+2.0$\,pp thing and $+1.1$\,pp boundary recall. Also, since our policy gradients allows our training to use hard sampled masks, $\lambda$ regulates the actual deployed token count rather than a soft proxy (Figure \ref{fig:lambda_control_ablation}).

\section{Sensitivity to Hyper-parameters}
\label{sec:supp_hparams}

A natural concern for an RL-based method is how sensitive it is to training hyper-parameters, and whether the added optimization introduces brittle tuning. To address this, we sweep each of SPLIT's PPO-specific and warm-start hyper-parameters, varying one at a time around the default configuration used in the main paper while holding all others fixed. 

\paragraph{PPO hyper-parameter sweeps}
Table~\ref{tab:hparam_single} varies the PPO clip ratio $\epsilon$, the number of rollouts $K$, the number of PPO epochs, the warm-start balancer rate, and the backbone learning rate during the warm-start. The results in Table~\ref{tab:hparam_single} correspond to the image classification task using the configuration: ViT-L, 224×224, 0.5 target token fraction. We observe that the accuracy is within $\pm0.2\%$ across all settings except in the case of (i) running $\geq 4$ PPO epochs over the same rollouts and (ii) backbone learning rates above $1\mathrm{e}{-6}$ degrade the pretrained features ($85.73\%$ at $1\mathrm{e}{-5}$). Both failure modes have well-understood causes: (i) over-optimizing the PPO objective and (ii) over-writing pretrained features. These are avoidable pitfalls and do not reflect instability unique to SPLIT.
 
\begin{table}[ht]
\centering
\caption{Single hyper-parameter sweeps on ImageNet-1K (ViT-L, $224\times224$,
target token fraction $0.5$). A run using the default hyper-parameters scores 86.49 at token fraction 0.586.}
\label{tab:hparam_single}
\begin{tabular}{lll}
\toprule
\textbf{Hyper-parameter (default)} & \textbf{Values swept} & \textbf{Test acc (\%)} \\
\midrule
PPO clip ratio $\epsilon$ ($0.10$)  & $0.05$ / $0.10$ / $0.20$ & 86.49 / 86.49 / 86.48 \\
Rollouts $K$ ($4$)                  & $2$ / $4$ / $6$          & 86.49 / 86.49 / 86.33 \\
PPO epochs ($2$)                    & $1$ / $2$ / $4$          & 86.50 / 86.49 / 83.28 \\
Warm-start balancer rate ($0.005$)  & $0.001$ / $0.005$ / $0.02$ & 86.44 / 86.49 / 86.54 \\
Backbone LR ($1\mathrm{e}{-6}$)     & $5\mathrm{e}{-7}$ / $1\mathrm{e}{-6}$ / $5\mathrm{e}{-6}$ / $1\mathrm{e}{-5}$ & 86.55 / 86.49 / 85.97 / 85.73 \\
\bottomrule
\end{tabular}
\end{table}

\paragraph{Dual rate and policy learning rate are coupled.} The Lagrangian dual rate $\eta_\lambda$ controls how quickly the multiplier $\lambda$ reacts to token-budget violations and the policy learning rate controls how fast the policy changes based on the combined cost weighted by $\lambda$. Due to this coupling of the two learning rates through $\lambda$, it makes sense to observe their joint behavior, which we present in Table~\ref{tab:hparam_dual} using the same configuration as Table~\ref{tab:hparam_single} with a target token fraction of 0.5 and varying only the two learning rate hyper-parameters.

We observe that an under-reactive dual ($\eta_\lambda = 1\mathrm{e}{-4}$) overshoots the target budget by $22$--$34\%$ as the cost does not adjust fast enough to enforce the budget. For $\eta_\lambda \geq 5\mathrm{e}{-4}$ the two hyper-parameters substitute for one another, anti-diagonal cells match within
$0.1\%$ and effectively behave as a single degree of freedom. 
 
\begin{table}[ht]
\centering
\caption{Joint sweep of the Lagrangian dual rate $\eta_\lambda$ and the RL-phase policy learning rate (ViT-L, $224\times224$, target token fraction $0.5$). Each cell reports test accuracy (\%) / realized token fraction. Defaults ($^{\dagger}$) are $\eta_\lambda = 5\mathrm{e}{-4}$ and policy LR $=5\mathrm{e}{-6}$.}
\label{tab:hparam_dual}
\begin{tabular}{lccc}
\toprule
 & \multicolumn{3}{c}{\textbf{Policy learning rate}} \\
\cmidrule(lr){2-4}
$\boldsymbol{\eta_\lambda}$ & $1\mathrm{e}{-6}$ & $5\mathrm{e}{-6}^{\dagger}$ & $1\mathrm{e}{-5}$ \\
\midrule
$1\mathrm{e}{-4}$            & 85.2 / 0.72 & 86.3 / 0.81 & 86.2 / 0.84 \\
$5\mathrm{e}{-4}^{\dagger}$  & 86.1 / 0.62 & 86.4 / 0.58 & 86.1 / 0.51 \\
$1\mathrm{e}{-3}$            & 86.3 / 0.55 & 86.1 / 0.52 & 86.1 / 0.51 \\
\bottomrule
\end{tabular}
\end{table}

\paragraph{Number of warm-start and RL iterations.} We also study how performance depends on the iterations used during the warm-start and RL phases of training through two complementary experiments. In the first experiment, we fix a $1500$-iteration warm-start on object detection and track performance across subsequent RL iterations. Table~\ref{tab:iter_det} shows that mAP increases monotonically with RL iterations ($58.06 \rightarrow 58.84$). The PPO phase contributes $+0.8$ mAP over the warm-start decisions at a matched token budget. In the second, we vary the split of iterations between the two phases on semantic segmentation while holding the \emph{total} iteration count fixed at $3000$. As Table~\ref{tab:iter_seg} shows, final mIoU stays within $0.05$ mIoU across all three splits and during training we observe that shorter warm-starts only enlarge the transient token-fraction oscillation early in RL. The warm-start therefore mainly stabilizes the beginning of RL rather than setting final quality. 

\begin{table}[ht]
\centering
\caption{RL-iterations ablation on object detection (EVA-02-B @ $1536$, $\rho = 0.70$) with a fixed $1500$-iteration warm-start. Default configuration marked $^{\dagger}$.}
\label{tab:iter_det}
\begin{tabular}{lcccc}
\toprule
\textbf{RL iterations} & $0$ (warm-start only) & $500$ & $1000$ & $1500^{\dagger}$ \\
\midrule
\textbf{COCO mAP} & 58.06 & 58.20 & 58.52 & 58.84 \\
\bottomrule
\end{tabular}
\end{table}
 
\begin{table}[ht]
\centering
\caption{Warm-start/RL iterations split on segmentation (EVA-02-L @ $640$, ADE20K, $\rho = 0.65$) with the total iteration count fixed at $3000$. Default configuration marked $^{\dagger}$.}
\label{tab:iter_seg}
\begin{tabular}{cccc}
\toprule
\textbf{Warm-start iters} & \textbf{RL iters} & \textbf{Total iters} & \textbf{Final mIoU} \\
\midrule
$1000$           & $2000$           & $3000$ & 59.34 \\
$1500^{\dagger}$ & $1500^{\dagger}$ & $3000$ & 59.39 \\
$2000$           & $1000$           & $3000$ & 59.38 \\
\bottomrule
\end{tabular}
\end{table}

\section{Robustness to Distribution Shift}
\label{sec:supp_robustness}

A desirable property of any adaptive tokenization method is that it must be robust to deviations from the data distribution on which the tokenization was trained or tuned. We run experiments to observe how the trained policy reacts to (i) change in distribution of test data and (ii) corruption of input images. We compare the behavior of SPLIT against APT \citep{choudhury2025apt} to show the benefit of using a learnable approach as compared to a heuristic driven approach.

\paragraph{Distribution shift.} We applied the ImageNet-trained policy to five datasets at 224×224 without retraining the policy: CIFAR-100, ImageNet-V2, -Sketch, -R, -A. For ImageNet $\rightarrow$ CIFAR-100 (resized), we learn a linear probe fit with the backbone frozen, since the label space of CIFAR-100 is different from that of ImageNet. For the four ImageNet variants, we use the same 1000-class head (frozen); specifically ImageNet-R/A are standard 200-class subsets, and are scored by restricting the logits to the subset. We compare with APT in two forms: \emph{fixed}-$\tau$ , a direct transfer condition, since recalibration needs target-domain data, and $\tau$ \emph{recalibrated} per dataset to SPLIT's token fraction for a matched-budget accuracy comparison.

We run the experiments in Table~\ref{tab:robust_shift} with a target token fraction of $0.85$ while training. SPLIT maintains a token fraction within $0.74-0.90$ across all test datasets and is at most $\sim\!0.11$ from its training target. Fixed-$\tau$ APT swings over $0.50-0.89$, overspending on CIFAR-100, collapsing to $0.50$ on ImageNet-Sketch. Even against per-dataset recalibrated APT, SPLIT is more accurate on the four ImageNet variants by $+0.14$ to $+0.77$, while staying within $0.1$\,pp of the dense baseline.

\begin{table}[ht]
\centering
\caption{Transfer of the ImageNet-trained SPLIT policy to five datasets with no retraining (ViT-L, $224\times224$; training target token fraction $0.85$). For SPLIT and fixed-$\tau$ APT we report accuracy with the
realized token fraction in parentheses. ``APT recal.'' tunes $\tau$ per dataset
to SPLIT's token fraction.}
\label{tab:robust_shift}
\begin{tabular}{lcccc}
\toprule
\textbf{Dataset} & \textbf{Dense} & \textbf{SPLIT (tok.\ frac.)} & \textbf{APT fixed-$\tau$ (tok.\ frac.)} & \textbf{APT recal.} \\
\midrule
CIFAR-100 (probe) & 89.87 & 89.26 (0.795) & 89.60 (0.831) & 89.49 \\
ImageNet-V2       & 79.13 & 79.07 (0.90)  & 78.93 (0.89)  & 78.93 \\
ImageNet-Sketch   & 61.14 & 61.22 (0.74)  & 56.82 (0.50)  & 60.73 \\
ImageNet-R        & 77.19 & 77.06 (0.83)  & 75.71 (0.74)  & 76.42 \\
ImageNet-A        & 64.45 & 64.37 (0.90)  & 63.28 (0.89)  & 63.60 \\
\bottomrule
\end{tabular}
\end{table}

\paragraph{Input corruption.} To isolate the behavior in a controlled setting, we add Gaussian pixel noise of increasing standard deviation $\sigma$ to the ImageNet test data and evaluate at $336\times336$ (ViT-L). In Table~\ref{tab:robust_noise} we start with SPLIT and APT having the same token fraction under zero noise. As noise increases the entropy of every patch, APT progressively splits more patches and its token fraction climbs from $0.53$ to $0.95$, while SPLIT holds its budget ($0.53 \rightarrow 0.59$). 

\begin{table}[ht]
\centering
\caption{Gaussian pixel noise of standard deviation $\sigma$ added to ImageNet test data (ViT-L @ $336\times336$). Each entry is test accuracy (\%) at the realized token fraction. APT $\tau$ is calibrated to match SPLIT's zero-noise token fraction.}
\label{tab:robust_noise}
\begin{tabular}{lcc}
\toprule
\textbf{Noise $\sigma$} & \textbf{SPLIT acc @ tok.\ frac.} & \textbf{APT acc @ tok.\ frac.} \\
\midrule
0.00 & 87.7 @ 0.53 & 87.5 @ 0.53 \\
0.05 & 86.5 @ 0.53 & 86.7 @ 0.62 \\
0.10 & 84.9 @ 0.54 & 85.3 @ 0.75 \\
0.15 & 83.0 @ 0.56 & 82.8 @ 0.92 \\
0.20 & 80.1 @ 0.59 & 79.3 @ 0.95 \\
\bottomrule
\end{tabular}
\end{table}

\section{Training Costs}

\label{sec:supp_training_cost}
 
We quantify SPLIT's training overhead for EVA-02-L at $640\times640$ on ADE20K, with two splitting stages (three patch scales). Table~\ref{tab:training_cost} uses identical hardware (single A100 GPU) and batch size, and a matched $3000$-iteration schedule for all rows and reports per-iteration and per-image latency alongside total wall-clock training time. Even though SPLIT has two phases of training (warmstart and PPO), the training budget is \emph{matched, not doubled}: SPLIT partitions APT's $3000$-iteration schedule into $1500$ warm-start and $1500$ PPO iterations. Because the backbone is frozen throughout the PPO phase, the backbone in SPLIT receives \emph{half} the updates APT does. In total, SPLIT trains only about $12$ minutes longer than APT, an overhead that comes almost entirely from the PPO phase.
 
\begin{table}[ht]
\centering
\caption{Training cost on ADE20K (EVA-02-L @ $640$, two splitting stages, 1$\times$A100 GPU).}
\label{tab:training_cost}
\begin{tabular}{lcccc}
\toprule
\textbf{Config} & \textbf{ms/iter} & \textbf{ms/img} & \textbf{Iters} & \textbf{Total} \\
\midrule
Vanilla (dense)  & 963  & 60  & 3000 & 48.2 min \\
APT finetune     & 809  & 51  & 3000 & 40.45 min \\
\midrule
SPLIT warm-start & 910  & 57  & 1500 & 22.8 min \\
SPLIT PPO phase  & 1186 & 74  & 1500 & 29.6 min \\
SPLIT total      & ---  & --- & 3000 & 52.4 min \\
\bottomrule
\end{tabular}
\end{table}
 
It is important to note that all methods, including heuristic driven ones like APT would require updating the backbone to adapt to the change in token distribution with a new tokenization scheme. The warm-start phase plays exactly this role for SPLIT, and in no configuration does SPLIT's backbone receive more updates than APT's. On classification, where recipes are epoch-based, the PPO phase adds one extra epoch with the backbone frozen ($\sim\!80$ minutes total on $2\times$A100 for the full two-phase run). SPLIT's gains are therefore attributable to policy learning, not additional finetuning.

Beyond training cost, this dependence on updating the backbone also constrains how an adaptive tokenizer can be used at inference. Changing the token budget at run time not only requires a tokenizer that allows for adjusting the budget but also needs the backbone to adapt to the new distribution of tokens for optimal performance. APT allows for budget control by adjusting the threshold $\tau$ to change the token budget, but since the backbone was trained with a different budget, the backbone cannot perform optimally unless we separately train a backbone for the new $\tau$. SPLIT in its current formulation does not support changing the token budget without re-training, this can be approximately achieved without training by shifting probabilities of the decision network, but this still does not address the problem of the backbone not being tuned for a new budget. We note this per-budget retraining requirement as a limitation (Section \ref{sec:conclusion}) and as a direction for future work. The ideal resolution would train the backbone with a randomly sampled $\rho$ and a tokenizer that accepts $\rho$ as an input on which decisions are conditioned, resulting in a single model taking a desired token fraction directly as input and removing the need for any per-budget adaptation.

\section{Quantitative Assessment of Semantic Alignment}
\label{sec:supp_semantic}
 
To quantitatively show that SPLIT's fine-token allocation is semantically aligned, we test whether the finest ($16\times16$) patches land on semantically important regions. We measure fine-token recall on ADE20K \texttt{val} against SceneParse150 ground-truth masks (unlabeled / not-in-150 pixels are ignored). For each image we build (1) a binary map of which $16\times16$ cells the policy keeps at fine resolution, and (2) a patch-level ground-truth label by pixel majority vote per cell; \emph{thing} patches are object-class cells, and \emph{boundary} patches are cells whose majority label differs from that of a 4-neighbor. Thing (resp.\ boundary) recall is the fraction of such patches that receive a fine token, i.e., of all thing (resp.\ boundary) patches, what share is allocated fine resolution. We compare Random, APT (entropy), and SPLIT at matched budgets ($\sim\!0.7$ token fraction) in Table~\ref{tab:semantic_recall_breakdown}.
 
\begin{table}[ht]
\centering
\caption{Fine-token recall on ADE20K \texttt{val} at a matched token budget
($\sim\!0.7$). \emph{Thing} recall is the fraction of object-class patches, and
\emph{boundary} recall the fraction of class-boundary patches, that receive a
fine ($16\times16$) token.}
\label{tab:semantic_recall_breakdown}
\begin{tabular}{lccc}
\toprule
\textbf{Method} & \textbf{Token fraction} & \textbf{Thing recall} & \textbf{Boundary recall} \\
\midrule
Random                  & 0.700 & 0.697 & 0.698 \\
APT (entropy)           & 0.719 & 0.808 & 0.833 \\
SPLIT, warm-start only  & 0.708 & 0.887 & 0.934 \\
SPLIT, full             & 0.714 & \textbf{0.907} & \textbf{0.945} \\
\bottomrule
\end{tabular}
\end{table}
 
As a sanity check, the Random baseline recalls objects and boundaries at approximately its token budget ($0.697$ and $0.698$ at a $0.700$ fraction), confirming that the metric is properly calibrated. At a matched budget, SPLIT allocates fine tokens to a substantially higher share of objects ($+9.9$\,pp over APT) and boundaries ($+11.2$\,pp), and the PPO phase further sharpens alignment beyond the warm-start policy ($+2.0$\,pp thing, $+1.1$\,pp boundary). This quantifies what our qualitative visualizations (Fig \ref{fig:split_vs_apt_patch_viz}) shows: the entropy heuristic spends its budget on high-entropy but semantically empty texture, whereas the learned policy reallocates it toward task-relevant structure.

\section{Decision Network Architecture}
\label{sec:supp_decision_arch}

We also ablate various different decision network architectures with the same parameter count to justify our choice of the cross attention architecture. In Table \ref{tab:router_architecture} we test our choice against using 2 layer MLPs with input as (i) only the embedding of the un-split parent patch or (ii) embedding of the un-split parent concatenated with embedding of all the child patches. We find that using the embeddings of both the parent and child embedding as inputs to the decision network improves performance, and our cross attention design outperforms using simple MLP layers for this. 

\begin{table}[ht]
\centering
\begin{tabular}{lccc}
\toprule
\textbf{Decision Network Architecture} & \textbf{Param Count} & \textbf{\% Tokens} & \textbf{Val Acc (\%)} \\
\midrule
$\mathrm{MLP}(x_{parent})$ & 1.115M & 89.59 & 87.69 \\
$\mathrm{MLP}(\mathrm{concat}([x_{parent}, x_{child}^{(1)}, \ldots , x_{child}^{(4)}]))$ & 1.116M & 90.08 & 87.74 \\
Cross Attention \ref{decision-module} & 1.115M & 89.53 & \textbf{87.81} \\
\bottomrule
\end{tabular}
\vspace{0.1in}
\caption{Comparison of different decision network architectures using ViT-L 224/14}
\label{tab:router_architecture}
\end{table}

\section{Qualitative Results}
\label{sec:supp_qualitative}

\begin{figure*}[t]
    \centering
    \includegraphics[width=\textwidth,height=0.85\textheight,keepaspectratio]{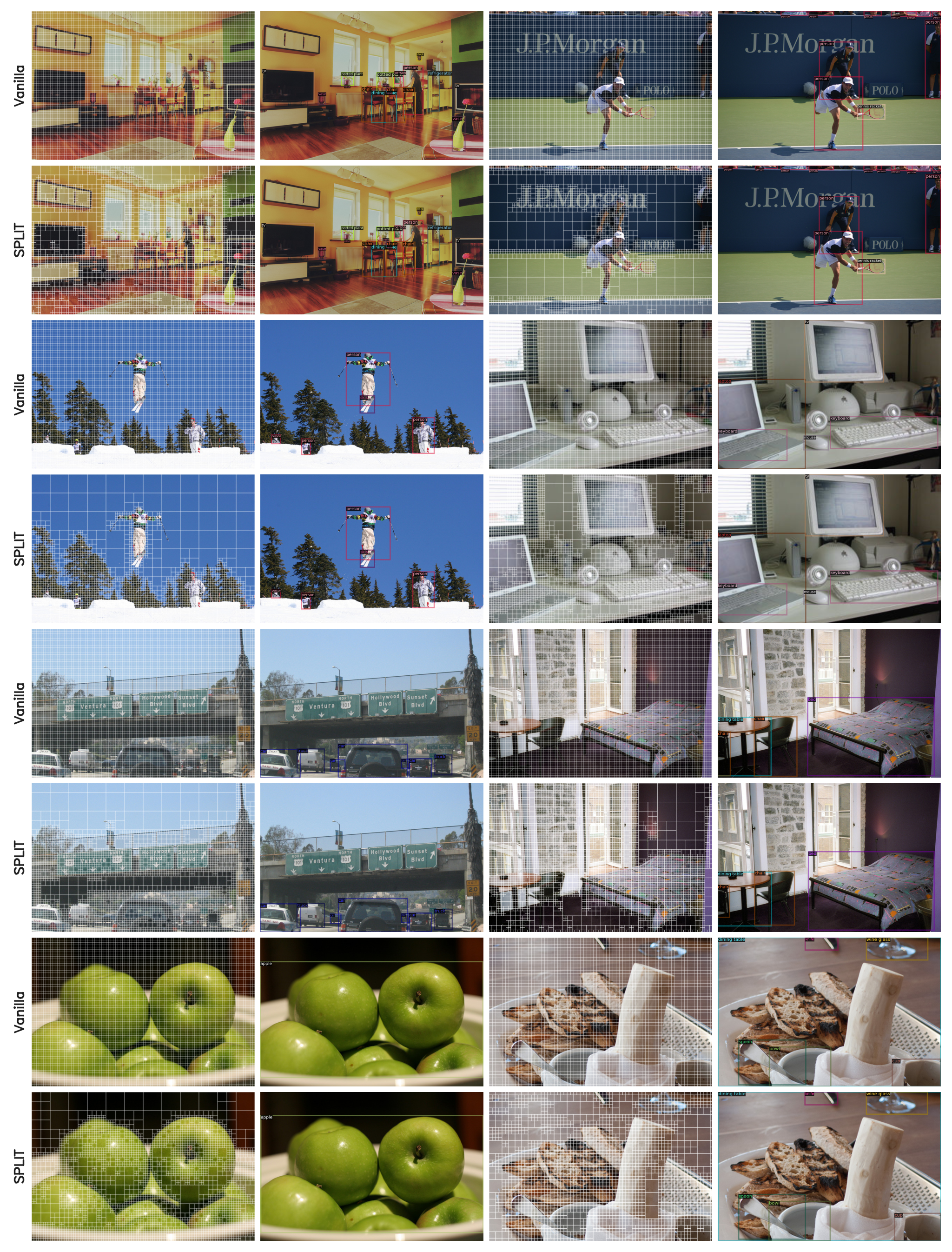}
    \caption{\textbf{Object detection qualitative comparison (Vanilla vs Ours).}
    We compare fixed-grid ViT inference (Vanilla) against our learned multi-stage patch splitting method.
    For each example, we visualize the patch assignment map and the corresponding detection output.
    Our method allocates finer patches to detail-heavy regions while preserving detection quality.}
    \label{fig:det_qualitative_vanilla_our}
\end{figure*}

\begin{figure*}[t]
    \centering
    \includegraphics[width=\textwidth,height=0.85\textheight,keepaspectratio]{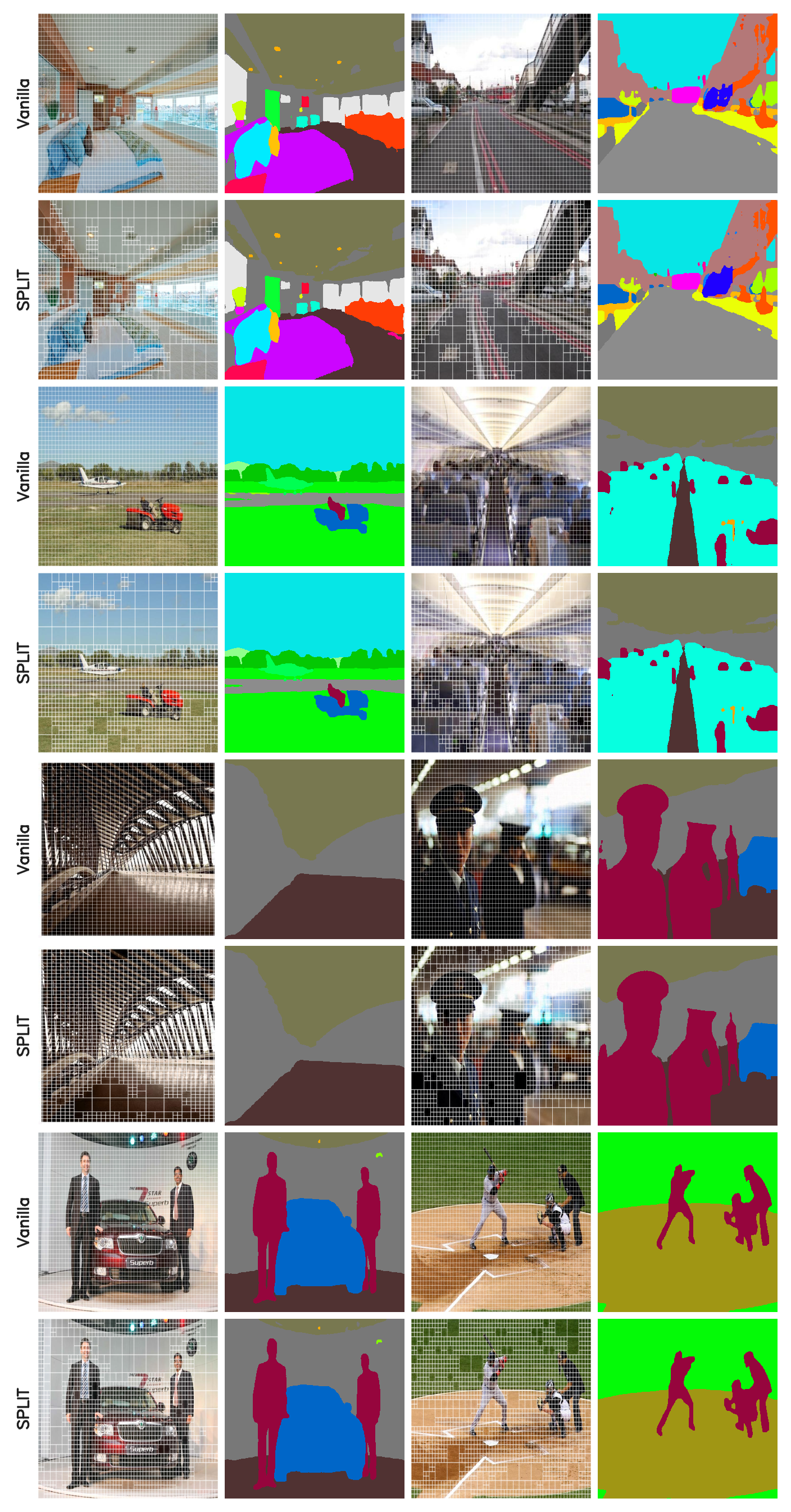}
    \caption{\textbf{Semantic segmentation qualitative comparison (Vanilla vs Ours).}
    We compare fixed-grid ViT inference (Vanilla) against our learned multi-stage patch splitting method.
    For each example, we visualize the patch assignment map and the corresponding segmentation output.
    Our method allocates finer patches to detail-heavy regions while preserving segmentation quality.}
    \label{fig:seg_qualitative_vanilla_our}
\end{figure*}

\begin{figure*}[t]
    \centering
    \includegraphics[width=\textwidth,height=0.85\textheight,keepaspectratio]{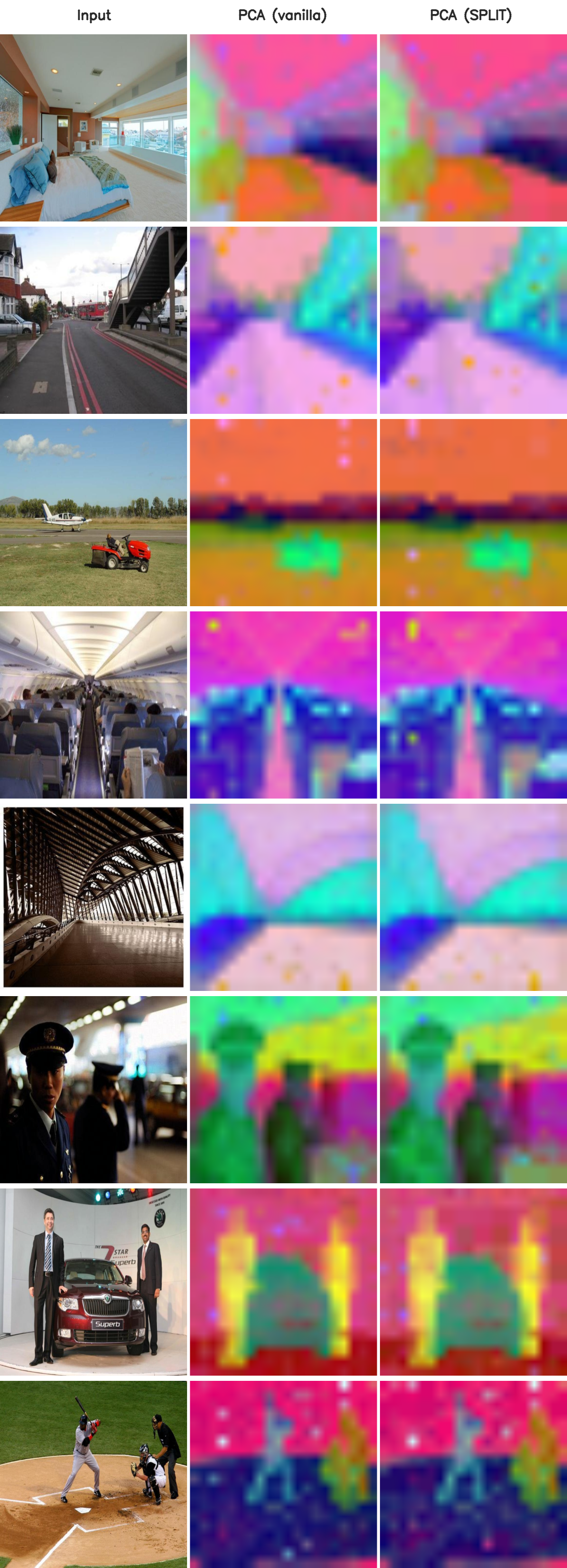}
    \caption{\textbf{PCA visualization of backbone feature maps (Vanilla vs Ours).}
    We project high-dimensional backbone features to RGB using a shared PCA basis fit on Vanilla features,
    then apply the same projection to our method.
    Comparable spatial structure in the PCA maps indicates that mixed-resolution splitting does not corrupt the learned representation.}
    \label{fig:pca_feature_comparison}
\end{figure*}

SPLIT generalizes effectively to downstream visual tasks that require
spatial precision, including object detection and semantic segmentation.
As illustrated in \Cref{fig:det_qualitative_vanilla_our} and
\Cref{fig:seg_qualitative_vanilla_our}, SPLIT reliably allocates larger
patches to uniform background regions while preserving fine-grained
structure with smaller patches around object boundaries and detail-heavy
areas. The resulting predictions match the vanilla full-resolution baseline
closely, supporting accurate bounding box regression and maintaining the
pixel-level fidelity required for segmentation. To verify that
mixed-resolution tokenization does not distort the model's internal
representations, \Cref{fig:pca_feature_comparison} compares backbone
feature maps under a shared PCA basis: SPLIT's features exhibit the same
spatial structure as the vanilla baseline, indicating that the multi-stage
splitting policy preserves the learned representation rather than
disrupting it.

\begin{figure*}[t]
    \centering
    \includegraphics[width=\textwidth,height=0.85\textheight,keepaspectratio]{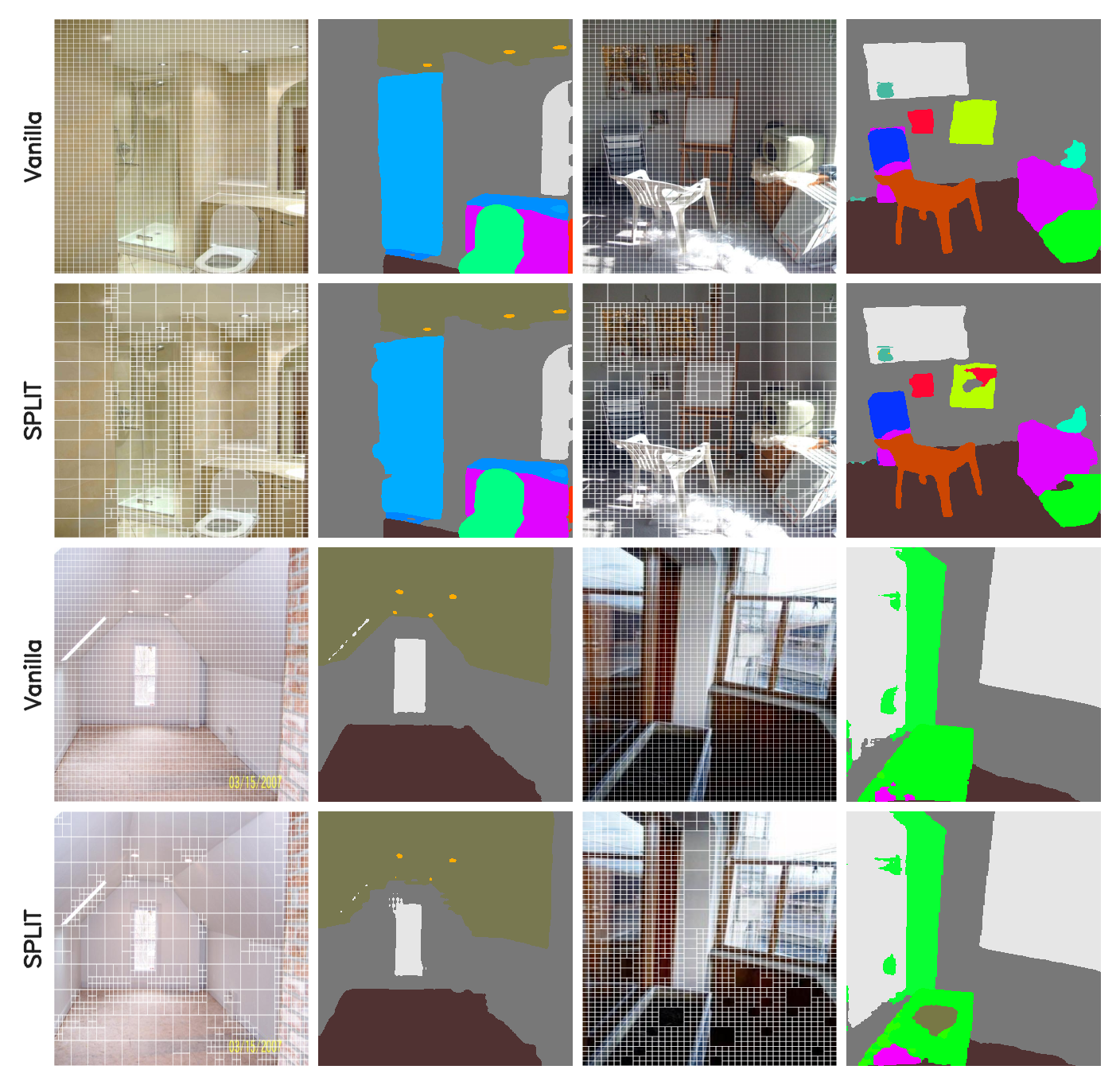}
    \caption{\textbf{Failure cases under aggressive token budgets.}
    When the target sequence length is pushed below the level used in
    our main experiments (here, $45\%$), SPLIT begins to under-tokenize semantically
    important regions, leading to coarse boundaries and missed thin
    structures in the predicted segmentation.}
    \label{fig:failure_seg}
\end{figure*}

\Cref{fig:failure_seg} illustrates the failure mode that emerges when the
token budget is pushed beyond the operating point used in our main
experiments. At a $45\%$ retention target, SPLIT begins to under-tokenize
genuinely informative regions, producing coarse object boundaries and
missing thin structures in the predicted segmentation. These cases mark
the boundary of the budget--accuracy trade-off: while SPLIT preserves task
performance across the budgets used in our main results, very aggressive
budgets force the policy to drop fine tokens in regions where they are
needed.

\begin{figure*}[t]
    \centering
    \includegraphics[width=\linewidth,height=0.85\textheight,keepaspectratio]{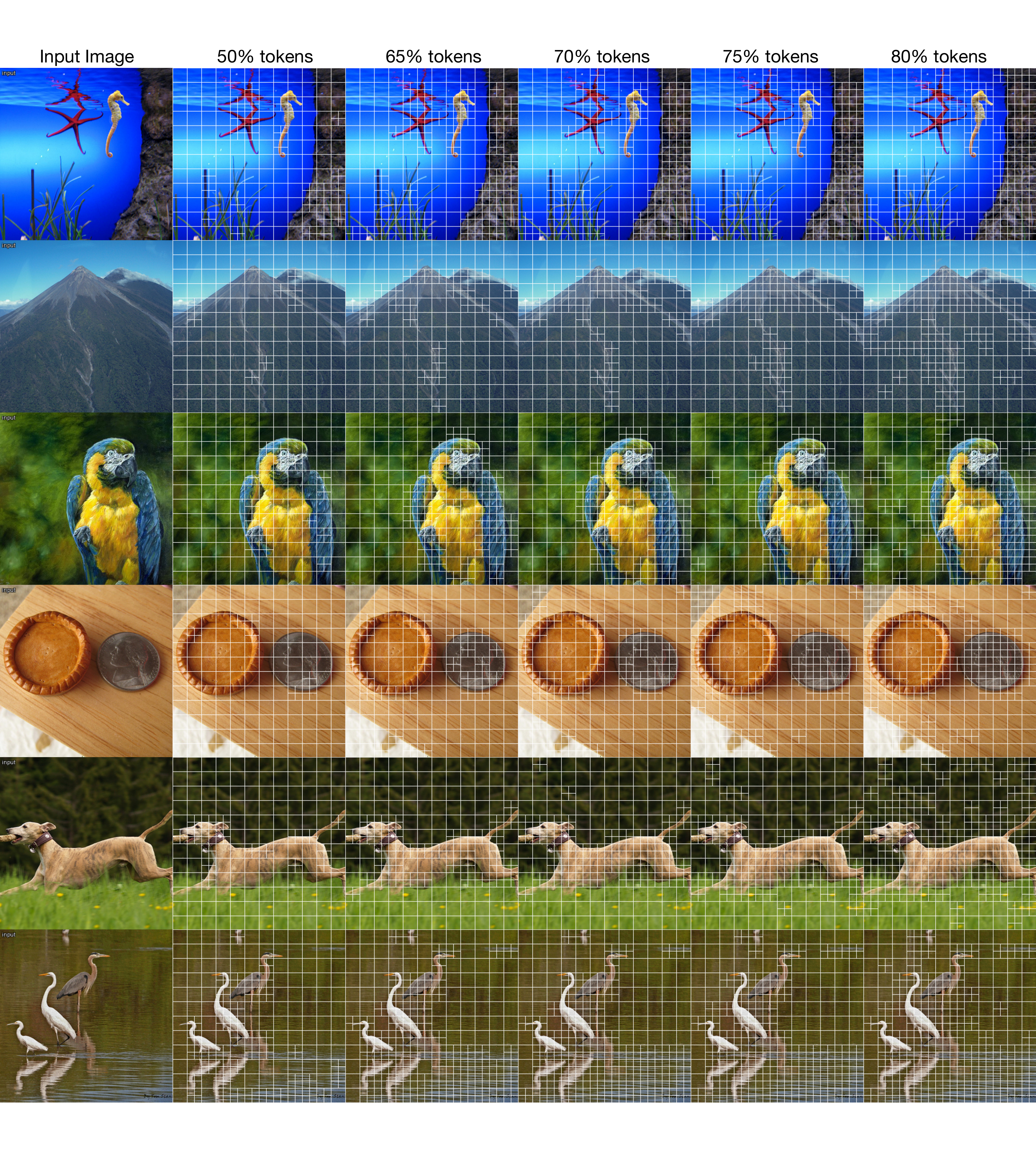}
    \caption{\textbf{Splitting decision at different token budgets} Visualizing the patch splitting decisions at various token budgets shows how the model progressively prioritizes patches based on content.}
    \label{fig:budget_sweep_viz}
\end{figure*}

\newpage

\end{document}